\documentclass[twocolumn]{stanford}
\usepackage{xcolor}

\usepackage{amsmath}

\usepackage[utf8]{inputenc} 
\usepackage[T1]{fontenc}    
\usepackage[hyphens]{url}
\usepackage[colorlinks=true,breaklinks=true]{hyperref}   
\usepackage{url}            
\usepackage{booktabs}       
\usepackage{amsfonts}       
\usepackage{nicefrac}       
\usepackage{microtype}      
\usepackage[table]{xcolor}         
\usepackage{xspace}
\usepackage{enumitem}
\usepackage{amsthm}
\usepackage{newtxtext,newtxmath}
\usepackage{subcaption} 
\usepackage{booktabs, multirow, makecell}
\usepackage{tcolorbox}
\tcbuselibrary{skins, listings, breakable}
\usepackage[T1]{fontenc}
\usepackage{longtable}
\usepackage{threeparttable}
\usepackage{tabularx}
\usepackage{pifont}
\usepackage{amsmath} 
\usepackage[utf8]{inputenc}
\usepackage{cleveref}

\usepackage{natbib}
\usepackage{wrapfig}

\usepackage{etoc}
\definecolor{darkblue}{rgb}{0,0.0,0.6}
\hypersetup{colorlinks=true, citecolor=darkblue, linkcolor=darkblue, urlcolor=darkblue}
\DeclareCaptionLabelFormat{subfig}{Figure~\thefigure\alph{subfigure}}
\setlist[itemize]{leftmargin=*, itemsep=2pt, topsep=0pt, parsep=0pt}
\setlist[enumerate]{leftmargin=*, itemsep=2pt, topsep=0pt, parsep=0pt}

\usepackage{multirow}
\usepackage{ragged2e}       

\definecolor{qback}{HTML}{F7FAFC}   
\definecolor{qframe}{HTML}{CBD5E1}  
\definecolor{aback}{HTML}{EFF6FF}   
\definecolor{aframe}{HTML}{93C5FD}  
\definecolor{jback}{HTML}{FFF7ED}   
\definecolor{jframe}{HTML}{FDBA74}  
\definecolor{okbg}{HTML}{DCFCE7}    
\definecolor{okfg}{HTML}{166534}
\definecolor{badbg}{HTML}{FEE2E2}   
\definecolor{badfg}{HTML}{7F1D1D}

\tcbset{
  breakable,
  sharp corners,
  boxrule=0.6pt,
  left=8pt,right=8pt,top=8pt,bottom=8pt,
  coltitle=black,
  fonttitle=\bfseries,
  before skip=8pt, after skip=10pt,
}

\crefname {promptbox}{Prompt box}{Prompt boxes}
\Crefname{promptbox}{Prompt~Box}{Prompt~Boxes}
\newtcolorbox{PromptBox}[1][]{
  colback=aback, colframe=aframe,
  title=Prompt,
  #1
}

\newtcolorbox{failurecase}[2]{%
  enhanced,
  colback=gray!3,
  colframe=gray!40,
  boxrule=0.8pt,
  arc=2mm,
  left=2mm,right=2mm,top=1.2mm,bottom=1.2mm,
  title=\textbf{Failure Case: #1} \hfill {\small #2},
  fonttitle=\normalsize,
}

\newtcolorbox{kagglecase}[2]{%
  enhanced,
  colback=gray!3,
  colframe=blue!15,
  boxrule=0.8pt,
  arc=2mm,
  left=2mm,right=2mm,top=1.2mm,bottom=1.2mm,
  title=\textbf{Kaggle Case Studies: #1} \hfill {\small #2},
  fonttitle=\normalsize,
}

\newtcolorbox{datasetexample}{
  breakable,
  colback=gray!5,
  colframe=black!70,
  boxrule=0.8pt,
  arc=4pt,
  left=6pt,
  right=6pt,
  top=6pt,
  bottom=6pt
}

\usepackage{listings}
\newcommand{\dsgym}{\raisebox{-.3em}{\rlap{{\textcolor{white}{ DSGym}}}\includegraphics[height=1.15em]{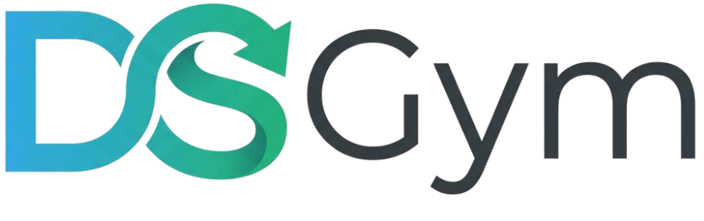}}\xspace}

\newcommand{\titlelogo}{\raisebox{-.3em}{\rlap{{\textcolor{metabg}{ DSGym}}}\includegraphics[height=1.15em]{figures/dsgym-logo.png}}\xspace}

\colorlet{rankFirst}{teal!28}
\colorlet{rankSecond}{teal!18}
\colorlet{rankThird}{teal!10}

\title{\titlelogo: A Holistic Framework for Evaluating and Training Data Science Agents}

\author[\spadesuit \heartsuit,*]{\text{Fan Nie}}
\author[\clubsuit,*]{\text{Junlin Wang}}
\author[\spadesuit,*]{\text{Harper Hua}}
\author[\heartsuit]{\text{Federico Bianchi}}
\author[\heartsuit]{\text{Yongchan Kwon}}
\author[\diamondsuit]{\text{Zhenting Qi}}
\author[\spadesuit]{\text{Owen Queen}}
\author[\heartsuit]{\text{Shang Zhu}}
\author[\spadesuit \heartsuit]{\text{James Zou}}
\affiliation[\spadesuit]{Stanford University}
\affiliation[\heartsuit]{Together AI}
\affiliation[\clubsuit]{Duke University}
\affiliation[\diamondsuit]{Harvard University}
\contribution{\textsuperscript{*}Equal Contribution}

\dataset{\url{https://huggingface.co/DSGym}}
\code{\url{https://github.com/fannie1208/DSGym}}
\metadata[Correspondence to]{\texttt{\{niefan,jamesz\}@stanford.edu}}

\begin{document}



\abstract{Data science agents promise to accelerate  discovery and insight-generation by turning data into executable analyses and findings. 
Yet existing data science benchmarks fall short due to fragmented evaluation interfaces that make cross-benchmark comparison difficult, narrow task coverage and a lack of rigorous data grounding. 
In particular, we show that a substantial portion of tasks in current benchmarks can be solved without using the actual data.
To address these limitations, we introduce \textsc{DSGym}, a standardized framework for evaluating and training data science agents in self-contained execution environments. 
Unlike static benchmarks, \textsc{DSGym} provides a modular architecture that makes it easy to add tasks, agent scaffolds, and tools, positioning it as a live, extensible testbed.
We curate \textsc{DSGym-Tasks}, a holistic task suite that standardizes and refines existing benchmarks via quality and shortcut solvability filtering. We further expand coverage with (1) \textsc{DSBio}: expert-derived bioinformatics tasks grounded in literature and (2) \textsc{DSPredict}: challenging prediction tasks spanning domains such as computer vision, molecular prediction, and single-cell perturbation.  Beyond evaluation, \textsc{DSGym} enables agent training via execution-verified data synthesis pipeline. As a case study, we build a 2,000-example training set and trained a 4B model in \textsc{DSGym}  that outperforms GPT-4o on standardized analysis benchmarks.
Overall, \textsc{DSGym} enables rigorous end-to-end measurement of whether agents can plan, implement, and validate data analyses in realistic scientific context.}



\twocolumn[{%
    \renewcommand\twocolumn[1][]{#1}%
    \maketitle
    \begin{center}
        \centering
      \includegraphics[width=0.95\textwidth]{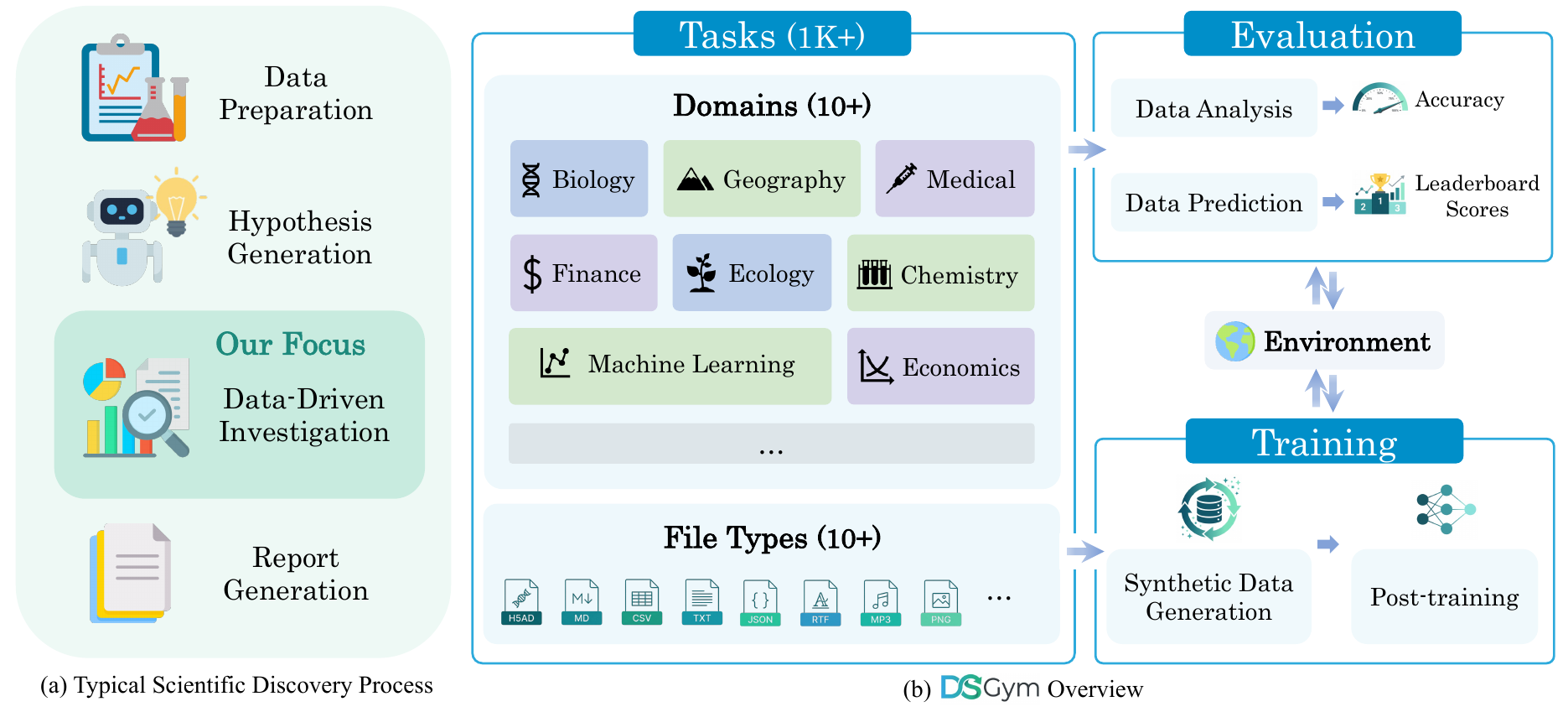}
        \captionof{figure}{(a) In the typical scientific discovery process, \textsc{DSGym} specifically focuses on the Data-Driven Investigation phase, where agents must bridge scientific hypotheses and empirical evidence through complex analysis. (b) We provide a unified environment spanning 10+ scientific domains and diverse file types. The framework enables a closed-loop ecosystem for both evaluation and training.} 
        \label{fig:main}
    \end{center}
    \vspace{0.0cm}  
}]

\etocdepthtag.toc{mtchapter}

\section{Introduction}

Data science serves as the computational engine of modern scientific discovery~\citep{scientific}. From identifying gene markers to predicting molecular properties, data science workflows turn datasets and scientific hypotheses (e.g., gene-disease associations) into empirical evidence. 
This process often requires heavy coding, intricate analysis, and tedious interactive computation~\citep{egg2025dabstep}, making it a natural target for Large Language Model (LLM) agents~\citep{Wang_2024} to automate these labor-intensive but structured tasks and substantially accelerate scientific progress~\citep{boiko2023emergent,chen2025largelanguagemodelbaseddata,Sun_2025}. Yet reliable automation demands a central requirement beyond textual reasoning: an agent’s decisions must be grounded in the data and validated by execution.

However, evaluating LLMs as data science agents remains challenging. The required skill set for data science is inherently broad, spanning iterative exploration, statistical inference, modeling, and domain-specific toolchains. Existing benchmarks can only capture fragments of this space, and they often differ in task formats, scoring conventions and execution environments~\citep{jing2024dsbenchfardatascience, majumder2024discoverybench, zhang2025datascibench, huang-etal-2024-da}. These inconsistencies make integration costly and hinder fair reproducible cross-benchmark comparison. 
More fundamentally, we revisit a core assumption underlying current data-science agent evaluation that file-grounded benchmarks (i.e., tasks accompanied by dataset files) necessarily measure data-dependent reasoning. We observe that a substantial portion of tasks in current file-grounded benchmarks can be solved even without accessing the files, revealing prompt-only shortcuts that inflate performance and confound measurement. Such shortcuts can arise from strong priors, pattern matching, or inadvertent contamination, undermining the validity of file-grounded evaluation as a proxy for genuine data interaction. In addition, current evaluations under-represent domain-specific scientific workflows, limiting our understanding of whether agents can support real scientific discovery rather than surface-level data manipulation.


To provide better support for the community, we propose \textsc{DSGym}, an integrated framework that unifies diverse data science evaluation suites behind a single API. We abstract the complexity of code execution behind containers that can be allocated in real time to execute code in a safe manner, allowing users to effectively run evaluations even on their local setups. Beyond providing a common execution layer, \textsc{DSGym} adopts a modular design that makes it straightforward to add new tasks, agent scaffolds, tools and evaluation scripts. This positions \textsc{DSGym} as a live, continuously extensible testbed for the community to measure and develop data science agents.

Beyond infrastructure, \textsc{DSGym} contributes \textsc{DSGym-Tasks}, a rigorously curated and expanded task ecosystem. 
We unify and audit widely used benchmarks under a standardized schema, and introduce a \emph{shortcut filtering} to remove tasks that can frequently be solved without data access. This yields a suite where performance more faithfully reflects data-dependent reasoning rather than prompt-only shortcuts. We further expand the evaluation scope by introducing two novel task suites: (i) \textsc{DSBio}: an expert-derived scientific analysis suite of 90 bioinformatics tasks grounded in academic literature, probing domain-specific scientific reasoning and tool use, and (ii) \textsc{DSPredict}: end-to-end modeling challenges sourced from recent Kaggle competitions spanning computer vision, molecular prediction, single-cell perturbation and so on, evaluating whether agents can build functional pipelines and iteratively improve predictive performance.

Using \textsc{DSGym}, we conduct a comprehensive study of frontier proprietary and open-weights LLMs, yielding detailed findings across general data analysis, domain-specific scientific workflows, and end-to-end modeling tasks. We find that even frontier models substantially underperform on scientific workflows: over 80\% of annotated failures are due to domain-grounding errors, such as misinterpreting domain concepts or using domain-specific libraries incorrectly. We further identify two recurring agent behaviors, \emph{simplicity bias} and lack of verification, that become especially damaging for realistic modeling tasks: on the hard split of \textsc{DSPredict}, the \textit{medal} rate is near 0\% even though the \textit{valid submission} rate exceeds 60\%, indicating that agents frequently stop after producing a runnable but under-optimized solution. Finally, although \textsc{DSGym} is primarily an evaluation framework, it can also support agent training. We demonstrate this by reusing \textsc{DSGym}'s agents and execution environments to generate execution-verified synthetic queries and trajectories, enabling a 4B model to reach competitive performance with frontier models such as GPT-4o on standardized analysis benchmarks. This highlights \textsc{DSGym}'s potential to function as both an evaluator and an active data factory.

\begin{figure*}[t]
    \centering
    \includegraphics[trim={3cm 1cm 3cm 2cm}, clip, width=\textwidth]{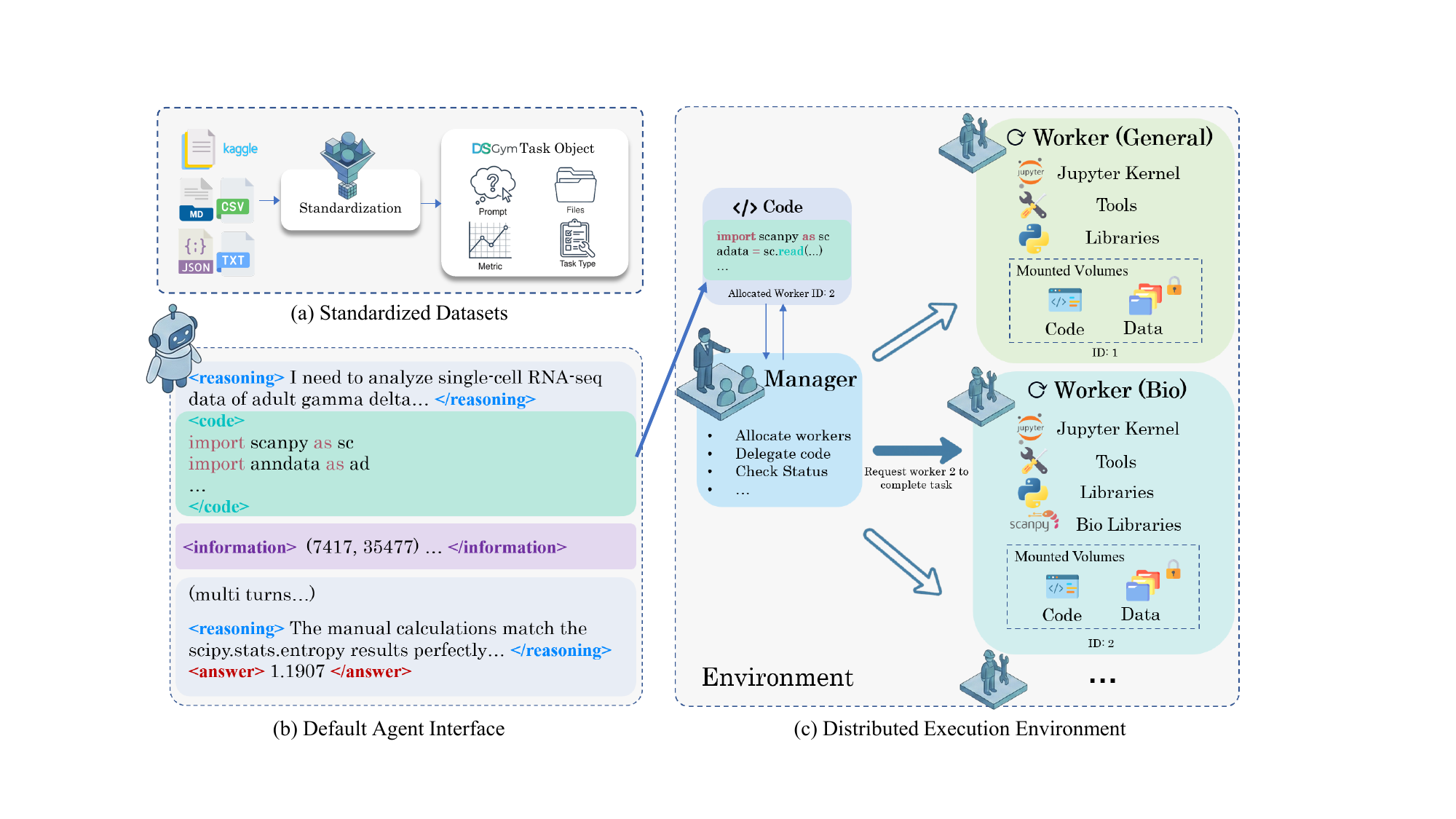}
    \caption{\textbf{The Architecture of \textsc{DSGym}.} 
        (a) \textbf{Standardized Tasks:} We aggregate heterogeneous data sources into a unified task object. (b) \textbf{Agent Interface:} \textsc{DSGym} provides a default CodeAct-like agent to interact with the environment. (c) \textbf{Execution Environment:} A central Manager container orchestrates the execution. Based on the task type, it dispatches agents to isolated Docker containers (Workers) pre-loaded with domain-specific libraries. Crucially, datasets are mounted as \textit{Read-Only Volumes}, while agents operate in a separate writable workspace. 
    }
    \label{fig:infra}
\end{figure*}

In summary, \textbf{our contributions} are as follows:
\begin{itemize}
    \item We show that existing data science benchmarks are vulnerable to shortcuts where agents can solve the task without using the actual data.
    \item We introduce \textsc{DSGym}, a unified, reproducible framework with standardized abstractions that enables cross-benchmark execution behind a single API. 
    \item We release \textsc{DSGym-Tasks}, a curated task ecosystem that standardizes and audits representative benchmarks, filters shortcut-solvable tasks, and expands coverage with \textsc{DSBio} and \textsc{DSPredict}.
    \item We benchmark frontier proprietary and open-weight LLMs on \textsc{DSGym} and analyze strengths and failure modes, revealing persistent gaps in domain-specific scientific workflows and common behaviors such as simplicity bias and insufficient verification.
    \item We demonstrate that \textsc{DSGym} enables execution-grounded trajectory synthesis for finetuning and we release a state-of-the-art small language model data science agent. 
\end{itemize}
\section{\textsc{DSGym}: A Unified Framework for Reproducible Data Science Agents}
\label{sec:dsgym}


Existing data science benchmarks evaluate useful but \emph{isolated} skills (e.g., statistical reasoning,
basic pandas/numpy usage) within heterogeneous execution environments,
making it difficult to assess agent abilities holistically or to compare results across benchmarks in a meaningful or reproducible way. \textsc{DSGym} addresses this gap by providing a unified, reproducible framework for executing and evaluating
data science agents across heterogeneous tasks and domains.

Rather than serving as another isolated benchmark, \textsc{DSGym} standardizes the representation of tasks, 
agent interfaces, and runtime environments. This infrastructure makes it possible to evaluate heterogeneous tasks, ranging from general-purpose data  analysis to scientific workflows and machine-learning modeling, under a single coherent protocol.
Our design is guided by three core requirements:

\textbf{(1) Realistic, data-dependent execution.}
Agents should be evaluated on tasks where correct solutions \emph{require} interacting with actual data files through
programmatic analysis.
This demands isolated execution environments, persistent state, controlled resource limits, and strict
filesystem separation to prevent contamination or unintended shortcuts.

\textbf{(2) Cross-benchmark standardization.}
To enable fair comparison across tasks from diverse domains,
\textsc{DSGym} standardizes task prompts, answer formats, evaluation metrics, and
environment definitions, removing inconsistencies arising from heterogeneous original benchmarks.

\textbf{(3) Modularity and extensibility.}
A modern testbed must support continuous growth rather than being a fixed static dataset.
\textsc{DSGym}'s modular design makes it easy to add new datasets, new evaluation scripts, new metrics, and new agent
scaffolds, keeping the benchmark up to date.
The same infrastructure also supports trajectory collection and synthetic data generation, enabling research on training data science agents (Section~\ref{sec:training}).

To operationalize this, we model the data science process as a standardized interaction loop: an \emph{Agent} perceives a problem defined by a \emph{Task} and executes code to solve it within a stateful, isolated \emph{Environment} (Fig.~\ref{fig:infra}). \textsc{DSGym} adopts a modular architecture consisting of these three components, and we discuss each component in detail below.


\subsection{Tasks and Datasets}
\label{subsec:datasets-abstraction}
\paragraph{Task Taxonomy.} We focus specifically on the \emph{data-driven investigation} phase of scientific discovery, where hypotheses are tested against empirical data through analysis and modeling.
To make this phase operational, \textsc{DSGym} organizes tasks into two categories:
\begin{enumerate}
    \item Data Prediction: A data prediction task provides a training dataset $D_\mathrm{train}$, a testing dataset $D_\mathrm{test}$ and a target metric $m$ (e.g., RMSE, accuracy). Agents should learn predictive models from $D_\mathrm{train}$ and make predictions on $D_\mathrm{test}$ to be evaluated by $m$.
    \item Data Analysis: A data analysis task provides one or more datasets $D=\{D_i\}$ and a question whose answer must be obtained by programmatic analysis of the data. Agents may employ any valid analytical procedure to reach the answer (e.g., statistical testing, causal inference, regression).
\end{enumerate}

Both categories require code execution over \emph{real data files}. We do not consider purely text-only QA as it does not assess data-dependent reasoning. We also do not consider visualization-centric tasks. Extending visualization-centric tasks is left for future work.

\paragraph{Unified Task Abstraction.} Regardless of category, \textsc{DSGym} expresses each task through a standardized \emph{Task Object}.  
Formally, a task instance is defined as a tuple
\((\mathcal{D}, \mathcal{P}, \mathcal{M}, \mathcal{Z}),\)

\begin{itemize}
    \item $\mathcal{D}$ denotes the data files required for execution (e.g., \texttt{.csv}, \texttt{.h5ad}).
    \item $\mathcal{P}$ specifies the query prompt, which standardizes the task instruction and background context.
    \item $\mathcal{M}$ defines the evaluation metric and any metric-specific configuration.
    \item $\mathcal{Z}$ contains structured metadata, such as the task category (analysis vs.\ modeling), 
          domain label, and keyword tags used for fine-grained capability analysis.
\end{itemize}

\paragraph{Dataset Organization.}
\textsc{DSGym} has a higher-level \textbf{Dataset} abstraction to manage the collections of tasks. A Dataset is a collection of individual tasks sharing common domains and evaluation protocols, easing the integration of new tasks and benchmarks. 


\subsection{Agents}

The Agent class acts as a wrapper around a base LLM and provides functionality for integrating various base models. 
It is also used for training language models as agents that take the history of all past actions and observations and return the next reasoning and action to take.

While users can integrate new agent architecture into \textsc{DSGym}, we provide a default agent interface as shown in Figure~\ref{fig:infra}(b). The agent interacts with the environment through a structured multi-turn loop. In each step, the agent outputs decision blocks in specific tags:
\begin{itemize}
    \item \texttt{<reasoning></reasoning>} for articulating analytical plans or reflecting on progress,
    \item \texttt{<code></code>} for writing executable Python code to perform analysis or call tools.
    \item \texttt{<answer></answer>} for submitting the final solution when the analysis is complete.
\end{itemize}
Executable outputs are captured by the environment and returned in \texttt{<information>} tags. This standardized interface ensures that models are evaluated on their reasoning and coding capabilities rather than their ability to parse arbitrary prompt formats.

\subsection{Environment}

\begin{figure*}[t]
    \centering
    \includegraphics[width=\textwidth]{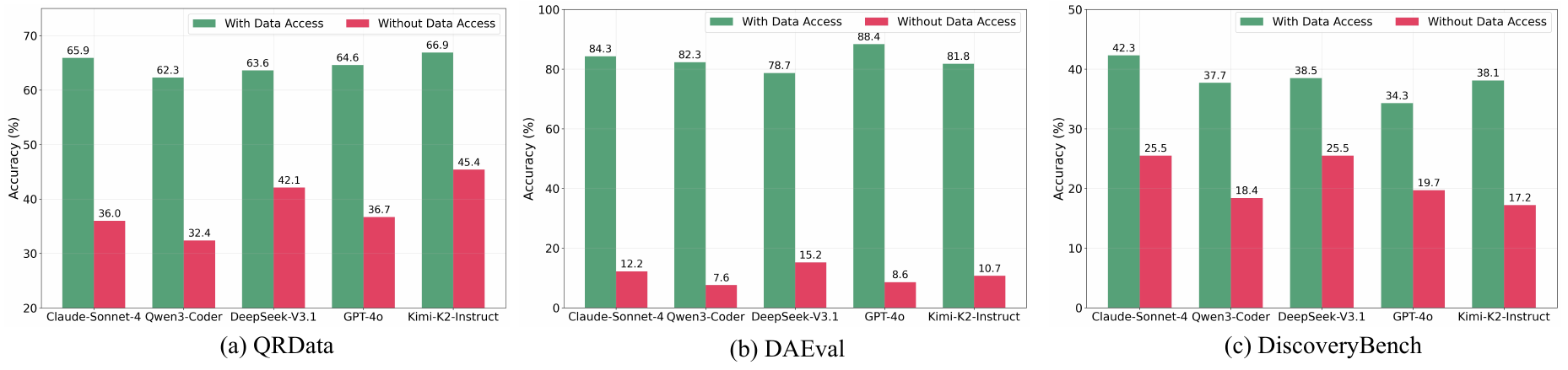}
    \caption{\textbf{Accuracy with or without data access on three file-grounded benchmarks.}
        We observe that even when real data files are not provided, agents can still answer a substantial fraction of questions correctly, suggesting that existing benchmarks can be partially solved via memorization, pattern matching, or priors rather than genuine data interaction.}
    \label{fig:dataset_motivation}
\end{figure*}

\begin{figure*}[t]
    \centering
    \includegraphics[width=\textwidth]{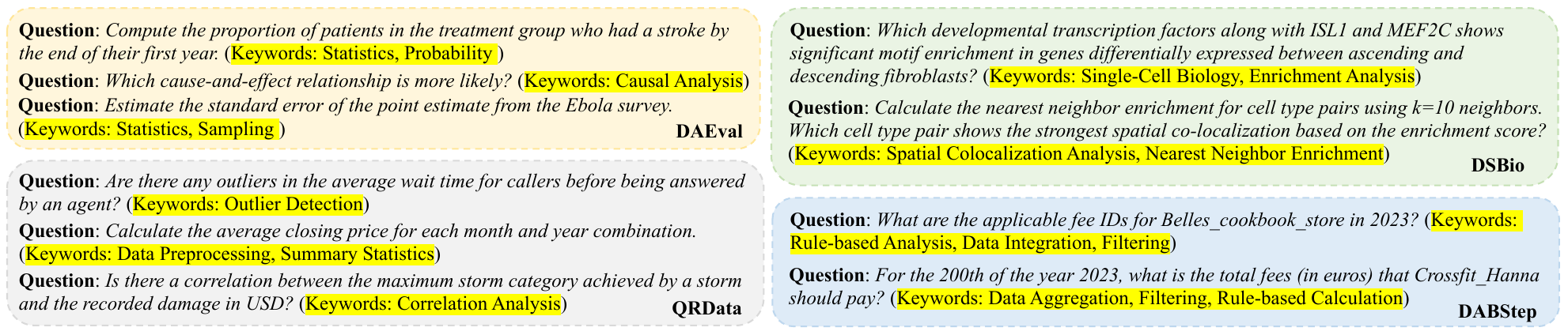}
    \caption{\textbf{Example questions across data science benchmarks.} Existing datasets such as \textsc{QRData}, \textsc{DAEval}, and \textsc{DABStep} mainly target general or applied data-science operations. \textsc{DSGym} complements these with new domain-specific scientific tasks (e.g., bioinformatics) that require specialized workflows and terminology.}
    \label{fig:dataset_motivation2}
\end{figure*}

Reproducible environments, controllable resources allocations and the availability of execution traces are all fundamental properties that a data science framework should offer.
In particular, data science workflows are inherently iterative and exploratory, demanding persistent memory state to efficiently manipulate large datasets. To support this, \textsc{DSGym} runs each agent trajectory inside a dedicated Jupyter kernel hosted within an isolated container. We adopt Jupyter due to its wide adoption in data science, but such a framework could be easily extended to RStudio or other environments.

The execution system follows a \emph{manager–worker} architecture for executing actions on separated environments. A central manager container orchestrates the entire system by allocating a fresh worker container at the start of each agent trajectory, binding read-only dataset mounts and writable workspace, and routing code requests. Each worker hosts an independent Jupyter kernel, ensuring complete isolation of Python environments, process state, and filesystem artifacts. Our execution system has the following features: 

\begin{itemize}
    \item \textbf{Stateful execution.} The environment preserves state across interaction steps: variables, trained models, and intermediate files generated in previous turns remain accessible in subsequent ones unless explicitly cleared. Resource limits on CPU, memory, and wall-clock time are enforced per container and can be user-specified. Once an agent submits an \texttt{<answer>} or generates a submission file, the artifacts are extracted and evaluated against the metric $\mathcal{M}$ in a clean, independent process. This ensures that the agent's environment state cannot interfere with the evaluation logic.
    \item \textbf{Tool integration.} The environment supports \emph{code-represented tools} that are functions callable from within the agent’s Python code and executed inside the kernel. In the current release, we include a lightweight web-search tool as an example, while users can easily register additional tools (e.g., domain-specific databases) without altering the core system.
    \item \textbf{Domain-specific containers.} Workers can be instantiated with different container images to accommodate domain-specific dependencies and tools. The manager automatically allocates each task to the appropriate container type, allowing heterogeneous tasks to execute within a unified infrastructure.
    \item \textbf{Filesystem Protection.} To ensure reproducibility and prevent invalid shortcuts, the environment enforces strict filesystem permissions. Dataset files are mounted to the container's volumes with \textbf{read-only} permissions. Agents operate in a separate, isolated writable workspace.
    \item \textbf{Environment Cycling.} Environments can be restarted and cycled so that users can effectively decide how many agents to run in parallel and they can define their own batching mechanisms.
\end{itemize}
This architecture enables \textsc{DSGym} to execute hundreds of trajectories in parallel while maintaining strict isolation, providing a scalable foundation for both evaluation in parallel and training of data science agents.

\section{Limitations of Existing Data Science Benchmarks}
\label{sec:benchmark-limitations}

Evaluating LLMs as data science agents requires moving beyond simple code generation to measuring execution-grounded reasoning and analysis pipelines over real-world datasets. While pioneering, existing benchmarks often fail to fully capture this process. Our audit of current benchmarks reveals three systemic limitations that hinder rigorous evaluation:

\textbf{Lack of Rigorous Data Grounding.} A core assumption of current file-grounded benchmarks is that tasks requiring associated data files necessarily measure data-dependent reasoning. However, our analysis reveals a pervasive ``shortcut" phenomenon: \textit{many questions can be answered correctly without reading the data.} As shown in \cref{fig:dataset_motivation}, across three prominent benchmarks, agents consistently achieve substantial accuracy even when data files are withheld. QRData shows only an average of only 40.5\% drop in performance across tasks while DAEval and DiscoveryBench show only 86.8\% and 44.4\% drops, respectively. 
    This suggests that performance is often inflated by data contamination, superficial pattern matching or domain priors rather than genuine interaction with the data.

\textbf{Task Invalidity and Inconsistency.} Several widely adopted benchmarks contain issues such as annotation errors, mismatched question–answer pairs, vague formatting instructions, or ambiguous multiple-choice options.  

\textbf{Limited Operation and Domain Coverage.} As illustrated in \cref{fig:dataset_motivation2}, existing benchmarks heavily overrepresent general statistics (e.g., descriptive statistics, aggregations or fitting small models), while providing limited coverage of domain-specific analytical workflows.
    Current agents are rarely tested on interpreting specialized terminology, processing raw scientific modalities (e.g., \texttt{.h5ad}), or utilizing domain-specific libraries. 

\section{\textsc{DSGym}-Tasks}

While Section~\ref{sec:dsgym} describes the unified architecture, we now turn to the \emph{dataset layer} of \textsc{DSGym}. \textsc{DSGym-Tasks} is designed to address the limitations identified in Section~\ref{sec:benchmark-limitations} and challenge agents across a \emph{diverse spectrum of data science tasks} that require interaction with real data files under a unified interface.
Our task suite spans both \emph{general} data science problems that represent the classic analysis workflows familiar to practitioners, and \emph{domain-specific scientific} tasks.
This diversity allows us to probe complementary dimensions of competence, including:
data manipulation, library proficiency,
strategic planning and domain grounded quantitative analysis. The curation of \textsc{DSGym-Tasks} are guided by three principles:

\begin{itemize}
    \item \textbf{Addressing flaws and inconsistencies in existing datasets.} We systematically audit and refine established datasets, removing invalid items and enforcing deterministic, reproducible answer formats to ensure reliability.
    \item \textbf{Enforcing genuine data interaction.} \textsc{DSGym} explicitly filters out tasks that remain solvable without real data access, restricting its scope to tasks where solutions are strictly \emph{data-dependent}.
    \item \textbf{Expanding operational and domain diversity.} We introduce expert-derived bioinformatics tasks and real-world end-to-end modeling challenges, stressing engineering competence and domain-grounded reasoning.
\end{itemize}

All tasks are executed in the containerized environment using the unified task abstraction (\cref{sec:dsgym}), ensuring fair, reproducible and holistic cross-domain evaluation.

\begin{figure}[ht!]
    \centering
    \includegraphics[width=0.9\linewidth]{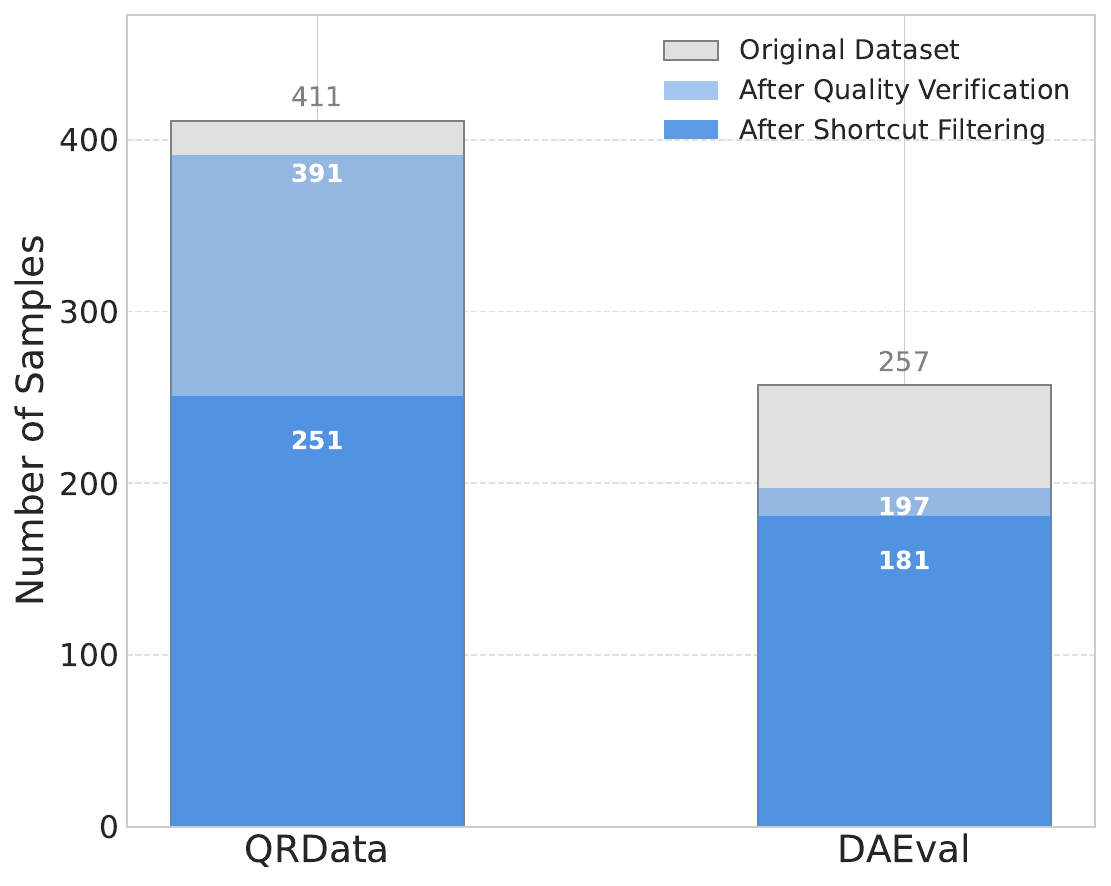}
    \caption{Filtering statistics after two-stage refinement.} 
    \label{fig:filter-statistics}
\end{figure}

\begin{figure*}[t]
    \centering
    \includegraphics[width=\textwidth]{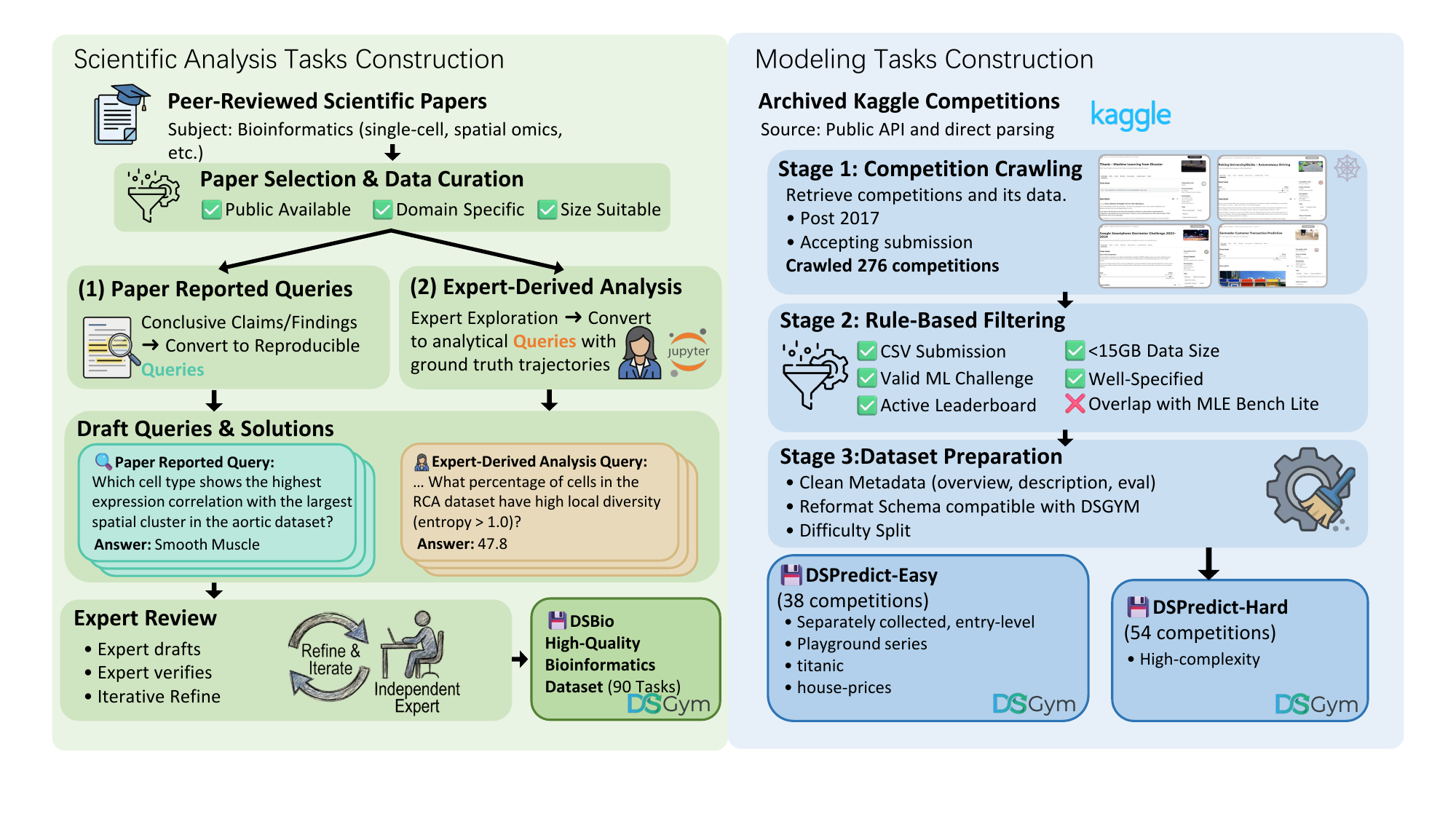}
    \caption{
        \textbf{Dataset Construction Pipeline.} 
        Our data construction pipeline curates domain-specific scientific tasks from academic literature and aggregates real-world predictive modeling challenges from Kaggle competitions.
    }
    \label{fig:dataset_pipeline}
\end{figure*}

\subsection{Refinement of Existing Datasets}

We begin by incorporating several widely used benchmarks into \textsc{DSGym} through two-stage refinement pipeline:
\begin{itemize}
    \item \textbf{Quality verification:}  
    We manually review every item, removing samples that are unscorable, ambiguous, or inconsistent with their gold answers.  
    Formatting issues (e.g., rounding precision, delimiter inconsistencies) are corrected to ensure deterministic evaluation.

    \item \textbf{Shortcut filtering:}  
    To operationalize data-dependence, we run five frontier LLMs on the remaining tasks \emph{without access to the data files}
    (\cref{fig:dataset_motivation}).
    If a majority ($\geq3$) of models still answer correctly, we mark the task as shortcut-solvable and exclude it from the final suite.
    This procedure filters out tasks frequently solvable without interacting with the provided data,
    including cases driven by memorization, domain priors, or surface-level heuristics,
    thereby retaining tasks that more directly require execution-grounded reasoning over data files.
\end{itemize}

Figure~\ref{fig:filter-statistics} shows the statistics of two-stage refinement. Below we summarize the refined subsets. 
More details are provided in Appendix~\ref{sec:refinement}.



\begin{itemize}
    \item \textbf{DAEval-Verified.} We remove samples lacking ground truths or containing misaligned question–answer pairs, refine answer-format guidelines (e.g., rounding precision inconsistencies) to match with the ground-truths, and correct typographical errors. The resulting dataset, categorized as data analysis tasks, provides short analytical queries that serve as a basic, general-purpose evaluation of data handling and statistical competence.
    \item \textbf{QRData-Verified.} We remove invalid multiple-choice queries with duplicate or ambiguous choices. This dataset focuses on statistical and causal reasoning over tabular data and belongs to the data analysis category.
    \item \textbf{DABStep.} DABStep comprises financial multi-step analytical queries that require reasoning across
    multiple data files. 
    \item \textbf{MLEBench-Lite.} We integrate \textsc{MLEBench-Lite} as a canonical data prediction benchmark within \textsc{DSGym}, ensuring full compatibility with our unified environment and metric registry.
\end{itemize}

\subsection{Scientific Analysis Tasks from Academic Literature}

To extend \textsc{DSGym} beyond generic data analysis, we curate \textsc{DSBio}, a new suite of
\textbf{90 bioinformatics tasks} derived from top-tier peer-reviewed publications and open-source scientific datasets.
We strategically select Bioinformatics as a \textit{pilot domain} to operationalize scientific discovery, as it uniquely combines high-dimensional, noisy data modalities that demand careful data inspection and domain-grounded statistical reasoning. These tasks probe critical dimensions of competence often underrepresented in existing benchmarks:
(1) interpreting unfamiliar data modalities (e.g., gene-expression matrices, spatial omics, high-dimensional noisy data),
(2) understanding domain-specific terminology and analytical conventions, and
(3) executing workflows with specialized libraries.
More details about \textsc{DSBio} are in Appendix \ref{appendix:dsbio}.


\paragraph{Task Construction Pipeline.} We select eight papers spanning single-cell omics, spatial omics, multi-omics integration, and human genetics. Papers are chosen only if they provided publicly available datasets of a size suitable for loading and analysis within our sandbox environment, avoiding excessive computational overhead. To ensure both coverage and depth, we construct tasks via two complementary ways:


\textbf{(1) Reproduction of Reported Findings.}  
We identify conclusive claims or quantitative findings reported in the original publications
and convert them into executable queries.  
To ensure compatibility with \textsc{DSGym}, a query is included only if:
(i) it can be answered purely from the provided dataset without visual inspection of figures,  
(ii) it produces a deterministic numerical or factual output.


\textbf{(2) Expert-Derived Follow-Up Analyses.}  Domain experts conduct a deep exploratory analysis of each dataset in a Jupyter notebook and design queries that require comprehensive, bottom-up reasoning from raw data, rather than simple information retrieval. We intentionally emphasized analytical difficulty by focusing on tasks involving statistical modeling, multi-dataset integration, and minimal reliance on pre-wrapped, domain-specific software packages. 




\paragraph{Iterative Expert Review.} To ensure the quality of the tasks and address the issue of nondeterminism common in scientific open-ended tasks, we implement an iterated expert verification process:
\begin{enumerate}
    \item A primary expert who drafts the analysis provides the task query and a `Gold Notebook' solution. 
    \item An independent expert reviews the quality and difficulty of the task and attempts to solve the task given only the prompt and data.
    \item If both solutions match and the task demonstrates sufficient analytical depth, it is accepted; otherwise, if the task is too simple, ambiguous or not deterministic, it is discarded or refined and re-reviewed until consensus is reached.
\end{enumerate}

\paragraph{Future Extensions to Other Scientific Domains}
While the current release focuses on bioinformatics, this construction pipeline is domain-agnostic and designed to extend to fields such as geoscience, computational chemistry and economics in future .

\subsection{Data Prediction Tasks from Kaggle Competition}

To capture realistic end-to-end modeling workflows, we implement a fully automated pipeline that continuously collects, filters, and standardizes completed Kaggle competitions into the \textsc{DSGym} format. 
The pipeline consists of three sequential stages: (i) competition crawling, (ii) rule-based filtering, and (iii) dataset preparation (~\cref{fig:dataset_pipeline}).

\paragraph{Stage 1: Competition Crawling.}
We first deploy a crawler that retrieves all archived Kaggle competitions through the Kaggle public API. 
For each competition, the crawler extracts the complete descriptions from web pages and the corresponding data files are automatically downloaded. Given the large number of available Kaggle competitions, we restrict our crawl to those that closed after 2017, and still accept submissions. In total, this stage collected 276 competitions spanning a broad range of challenges across structured data, text, and image modalities.

\paragraph{Stage 2: Rule-Based Filtering.}
Next, we apply a rule-based filtering to ensure that only well-structured, executable data science competitions remain:
\begin{enumerate}
    \item Format \& Size: Submissions must be in CSV format; datasets must be under 15 GB for hardware feasibility.
    \item Core Focus: Must be a valid ML challenge (no CTFs or code golf) requiring meaningful engineering and pipeline design.
    \item Evaluation: Requires an active leaderboard for quantitative benchmarking.
    \item Clarity: Objectives and data structures must be well-specified to support reproducibility.
    \item Uniqueness: Minimal  overlap with MLE Bench Lite.
\end{enumerate}

A complete rule set is deferred to \cref{appendix:kaggle:filter}. 

\paragraph{Stage 3: Dataset Preparation.}
For each filtered competition, we perform standardized data preparation.
Competition metadata (overview, data description, and evaluation details) are cleaned and reformatted into a consistent schema compatible with the \textsc{DSGym} dataset abstraction (\cref{subsec:datasets-abstraction}). 
We further categorize the resulting competitions into two difficulty splits.

\textbf{DSPredict-Easy.} The \textsc{DSPredict-Easy} split consists primarily of competitions from the Kaggle \emph{Playground Series} and two canonical introductory datasets—\textit{Titanic: Machine Learning from Disaster} and \textit{House Prices: Advanced Regression Techniques}.  These challenges are intentionally simple in both data structure and task objectives, making these \textbf{38 competitions} ideal testbeds and entry points for data science experimentation.

\textbf{DSPredict-Hard.} The high-complexity challenges we filtered earlier form the \textsc{DSPredict-Hard} split. After all filtering and validation, this split includes \textbf{54 competitions}. For challenges with multiple stages, we consistently used the second stage, as the official leaderboard metrics correspond to that stage. The final dataset suite preserves the original leaderboard metric definitions while ensuring full compatibility with \textsc{DSGym} containers and evaluation tools.


\begin{figure}
    \centering
    \includegraphics[width=0.9\linewidth]{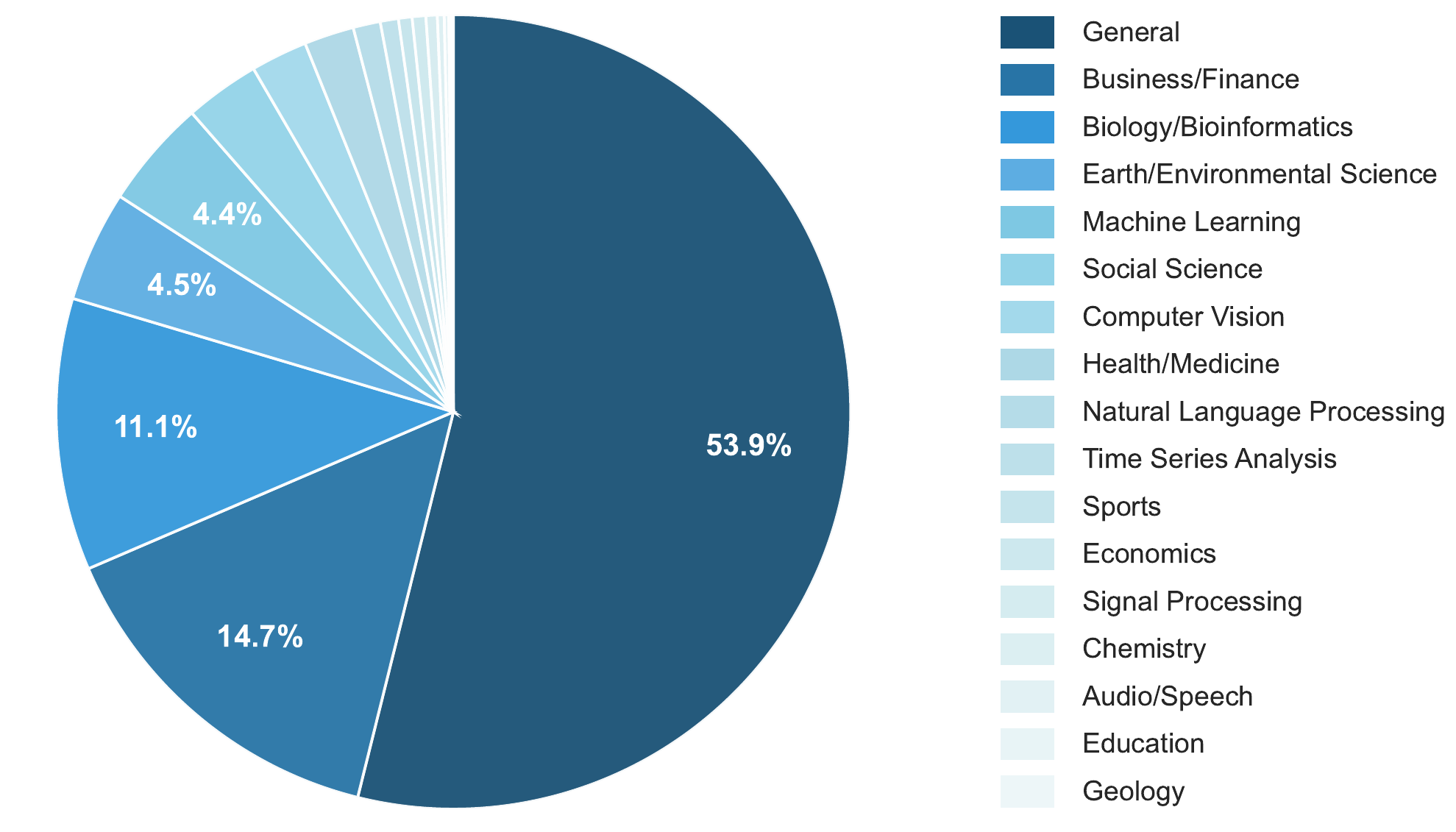}
    \caption{Percentage of task domains}
    \label{fig:domain}
\end{figure}

\begin{table}[!t]
    \centering
    \begin{tabular}{c|cc}
    \toprule
        Type & Data Analysis & Data Prediction\\
    \midrule
        Number & 972 & 114 \\
    \bottomrule
    \end{tabular}
    \caption{Statistics of \textsc{DSGym-Tasks}.}
    \label{tab:task_statistics}
\end{table}

\begin{table*}[!ht]
\centering\small
\resizebox{0.85\textwidth}{!}{
\begin{tabular}{lcccc}
\toprule
\textbf{Model} & \textbf{QRData-Verified} (\%) & \textbf{DABStep-easy} (\%) & \textbf{DABStep-hard} (\%) & \textbf{DAEval-Verified} (\%) \\
\midrule
\multicolumn{5}{c}{\textbf{Proprietary Models}} \\
\midrule
GPT-5.1 (high) & 60.16 & 73.61 & 13.23 & 89.50 \\
GPT-5.1 (none) & 58.96 & 70.83 & 11.9 & 87.85 \\
GPT-5 (medium) & \cellcolor{rankSecond}61.75 & 75.00 & 28.31 & 89.50 \\
GPT-4o & 60.24 & 73.61 & 7.41 & \cellcolor{rankSecond}92.26 \\
Claude Sonnet 4.5 & \cellcolor{rankThird}61.35 & \cellcolor{rankFirst}\textbf{83.33} & \cellcolor{rankFirst}\textbf{37.04} & \cellcolor{rankThird}91.71\\
Claude Sonnet 4 & 59.06 & \cellcolor{rankSecond}81.94 & \cellcolor{rankSecond}31.75 & 90.91  \\
\midrule
\multicolumn{5}{c}{\textbf{Open-sourced Models}} \\
\midrule
Qwen3 235B Instruct & 54.18 & 73.61 & 17.46 & 85.08 \\
Qwen3-Coder 480B & 54.72 & 75.00 & 14.29 & 90.61 \\
Kimi K2 Instruct & \cellcolor{rankFirst}\textbf{63.68} & \cellcolor{rankThird}77.78 & \cellcolor{rankThird}28.84 & \cellcolor{rankFirst}\textbf{92.82}  \\
GPT-OSS-120B & 47.95 & 70.83 & 7.94 & 84.53  \\
Deepseek-v3.1 & 57.37 & 76.39 & 21.96 & 82.32\\
\midrule
Qwen2.5-7B-Instruct & 35.04 & 47.22 & 2.38 & 50.56\\
Qwen3-4B-Instruct & 45.27 & 58.33 & 2.9 & 64.47 \\
\bottomrule
\end{tabular}}
\vspace{-2mm}
\caption{Accuracy performance comparison across standardized general data analysis datasets. 
}
\label{tab:data-analysis}
\vspace{-2mm}
\end{table*}

\paragraph{Distinction of DSPredict-Hard from MLE-Bench.}
The primary distinction between our curated \textsc{DSPredict-Hard} collection and \textsc{MLE-Bench Lite} lies in the recency and accessibility of the included competitions. Our dataset focuses on newer Kaggle challenges. The oldest from 2017 and several from 2024–2025, thereby reducing the likelihood of data leaks and ensuring that tasks better reflect contemporary machine learning and data science practices. In addition, we include only competitions that still accept submissions on Kaggle, allowing us to obtain official leaderboard scores.
This ensures accurate, up-to-date evaluation and maintains the benchmark’s relevance to current ML workflows.


\subsection{Dataset Statistics}

We show the domain distribution and task statistics of \textsc{DSGym-Tasks} in \cref{fig:domain} and \cref{tab:task_statistics}. 


\section{Evaluation}
\label{sec:experiments}

We conduct extensive experiments to evaluate state-of-the-art LLMs on \textsc{DSGym}. Our evaluation aims to answer: How do frontier models perform across the distinct capabilities of general data analysis, specialized scientific discovery, and end-to-end modeling?

\subsection{Evaluation Setup}
\textbf{Models.} We evaluate a suite of closed-source models (GPT-5.1, GPT-5, GPT-4o, \textsc{Claude Sonnet 4.5}, \textsc{Claude Sonnet 4}), open-weights models (\textsc{Qwen3-Coder 480B}, \textsc{Qwen3 235B Instruct}, \textsc{GPT-OSS-120B}, \textsc{DeepSeek-V3.1}, \textsc{Kimi-K2-Instruct}) and small models (\textsc{Qwen2.5-7B-Instruct}, \textsc{Qwen3-4B-Instruct}). Unless otherwise specified, all models are evaluated using the default \textbf{CodeAct} agent provided in \textsc{DSGym} with temperature $T=0$. Although \textsc{DSGym} environment supports tool integration (e.g., web search), all tools are disabled in all evaluations..

\textbf{Metrics.} For analysis tasks, we report exact-match accuracy with a slight numerical tolerance. For prediction tasks (e.g., MLEBench-Lite, \textsc{DSPredict}), we employ competition-specific leaderboards to derive three metrics: Valid Submission Rate, Above Median Rate, Any Medal Rate. For \textsc{DSPredict-Easy}, where medal rates are uninformative due to leaderboard saturation,
we report Percentile rank instead.
More details can be found in Appendix.~\ref{sec:experiment_detail}.

\subsection{Evaluation Results}

\begin{table*}[t]
\centering\small
\resizebox{0.8\textwidth}{!}{
\begin{tabular}{lcccc}
\toprule
\textbf{Model} & \textbf{Overall} (\%)& \textbf{Single-Cell Biology} (\%) & \textbf{Genetics} (\%) & \textbf{Spatial Transcriptomics
} (\%) \\
\midrule
\multicolumn{5}{c}{Closed-sourced Models} \\
\midrule
GPT-5.1 (high) & 38.89 & 43.10 & 28.57 & 36.36\\
GPT-5.1 (none) & 37.78 & 36.21 & 33.33 & 54.55 \\
GPT-5 (medium) & 32.22 & 34.48 & 33.33 & 18.18 \\
GPT-4o & 33.33 & 41.38 & 4.76 & 45.45 \\
Claude Sonnet 4.5 & \cellcolor{rankSecond}42.22  & 44.83 & 33.33 & 45.45 \\
Claude Sonnet 4 & 36.67 & 37.93 & 33.33 & 36.36 \\
\midrule
\multicolumn{5}{c}{Open-sourced Models} \\
\midrule
Qwen3 235B Instruct & 38.89 & 41.38 & 42.86 & 18.18\\
Qwen3-Coder 480B & 34.44 & 36.21 & 28.57 & 36.36 \\
Kimi K2 Instruct & \cellcolor{rankFirst}\textbf{43.33} & 44.83 & 42.86 & 36.36\\
GPT-OSS-120B & 25.56 & 27.59 & 14.29 & 36.36\\
Deepseek-v3.1 & \cellcolor{rankThird}40.00 & 41.38 & 38.10 & 36.36\\
Qwen2.5-7B-Instruct & 4.44 & 6.35 & 4.34 & 0 \\
Qwen3-4B-Instruct &  6.67 & 8.47 & 4.76 & 0 \\
\bottomrule
\end{tabular}}
\vspace{-2mm}
\caption{Accuracy Performance comparison on \textsc{DSBio} tasks.}
\label{tab:scienceagent}
\vspace{-4mm}
\end{table*}

\begin{table*}[t]
\centering\small
\resizebox{0.9\textwidth}{!}{
\begin{tabular}{l ccc ccc ccc}
\toprule
& \multicolumn{3}{c}{\textbf{MLEBench-lite}} & \multicolumn{3}{c}{\textbf{DSPredict-Hard} (Private)} & \multicolumn{3}{c}{\textbf{DSPredict-Easy} (Private)} \\
\cmidrule(lr){2-4} \cmidrule(lr){5-7} \cmidrule(lr){8-10}
\textbf{Model} & \textbf{Valid} & \textbf{Medal} & \textbf{Median} & \textbf{Valid} & \textbf{Medal} & \textbf{Median} & \textbf{Valid} & \textbf{Percentile} & \textbf{Median} \\
\midrule
GPT-5.1 (high) & \cellcolor{rankSecond}90.91 & \cellcolor{rankFirst}\textbf{22.73} & \cellcolor{rankFirst}\textbf{45.45} & \cellcolor{rankFirst}  \textbf{85.7} & \cellcolor{rankFirst} \textbf{4.8} & \cellcolor{rankFirst} \textbf{14.3} & \cellcolor{rankFirst}\textbf{100} & \cellcolor{rankFirst}\textbf{60.4} & \cellcolor{rankFirst}\textbf{75} \\
GPT-5.1 (medium) & \cellcolor{rankSecond}90.91 & \cellcolor{rankFirst}\textbf{22.73} & 31.82 & \cellcolor{rankSecond}  81.0 & \cellcolor{rankFirst}  \textbf{4.8} & 7.1 & 91.7 & \cellcolor{rankSecond}55.7 & \cellcolor{rankSecond}63.9 \\
GPT-5.1 (none) & 72.72 & 13.64 & 22.73 & 69.0 & 2.4 & \cellcolor{rankSecond} 10.3 & 97.2 & 45.7 & 41.7 \\
GPT-5 (medium) & 77.27 & 9.09 & 27.27 & 52.4 & 0 & 2.4 & 75 & 53.5 & 52.8 \\
Claude Sonnet 4.5 & 86.36 & \cellcolor{rankFirst}\textbf{22.73} & \cellcolor{rankSecond}36.36 & 71.4 & 0 & 4.8 & \cellcolor{rankFirst}\textbf{100} & 49 & 52.8 \\
Claude Sonnet 4 & \cellcolor{rankSecond}90.9 & 13.63 & 22.73 & \cellcolor{rankFirst}  \textbf{85.7} & 2.4 & 4.8 & \cellcolor{rankFirst}\textbf{100} & 44.4 & 36.1 \\
Qwen3-Coder 480B & \cellcolor{rankFirst}\textbf{100.0} & 9.09 & 22.72 & 66.7 & 2.4 & 5.9 & 86.5 & 42.9 & 33.3 \\
Qwen3 235B Instruct & 81.82 & 4.55 & 13.64 & 64.3 & 2.4 & 2.4 & 97.2 & 42.9 & 33.3 \\
Kimi K2 Instruct & 86.37 & 13.64 & 27.27 & 69 & 0 & 0 & 97.2 & 43.9 & 41.7 \\
Deepseek v3.1 & 86.37 & 13.64 & 27.27 & 76.2 & 2.4 & 7.1 & 86.8 & 30.9 & 7.9 \\
Qwen3-4B-Instruct & 50.0 & 4.55 & 9.09 & 40.5 & 0 & 0 & 67.6 & 28.2 & 5.4 \\
Qwen2.5-7B-Instruct & 0 & 0 & 0 & 4.8 & 0 & 0 & 35.1 & 17.5 & 0 \\
\bottomrule
\end{tabular}}
\vspace{-2mm}
\caption{Performance comparison across data prediction tasks. For \textsc{DSPredict}, results are reported on private test set. 
We report Valid Submission Rate (\textit{Valid}), Above Median Rate (\textit{Median}), and Any Medal Rate (\textit{Medal});
for \textsc{DSPredict-Easy}, we report Percentile rank instead of Medal. Details of metrics are deferred to Appendix~\ref{sec:experiment_detail}.
}
\label{tab:kaggle-analysis}
\vspace{-3mm}
\end{table*}

\begin{figure*}[t]
    \centering
    \includegraphics[width=\textwidth]{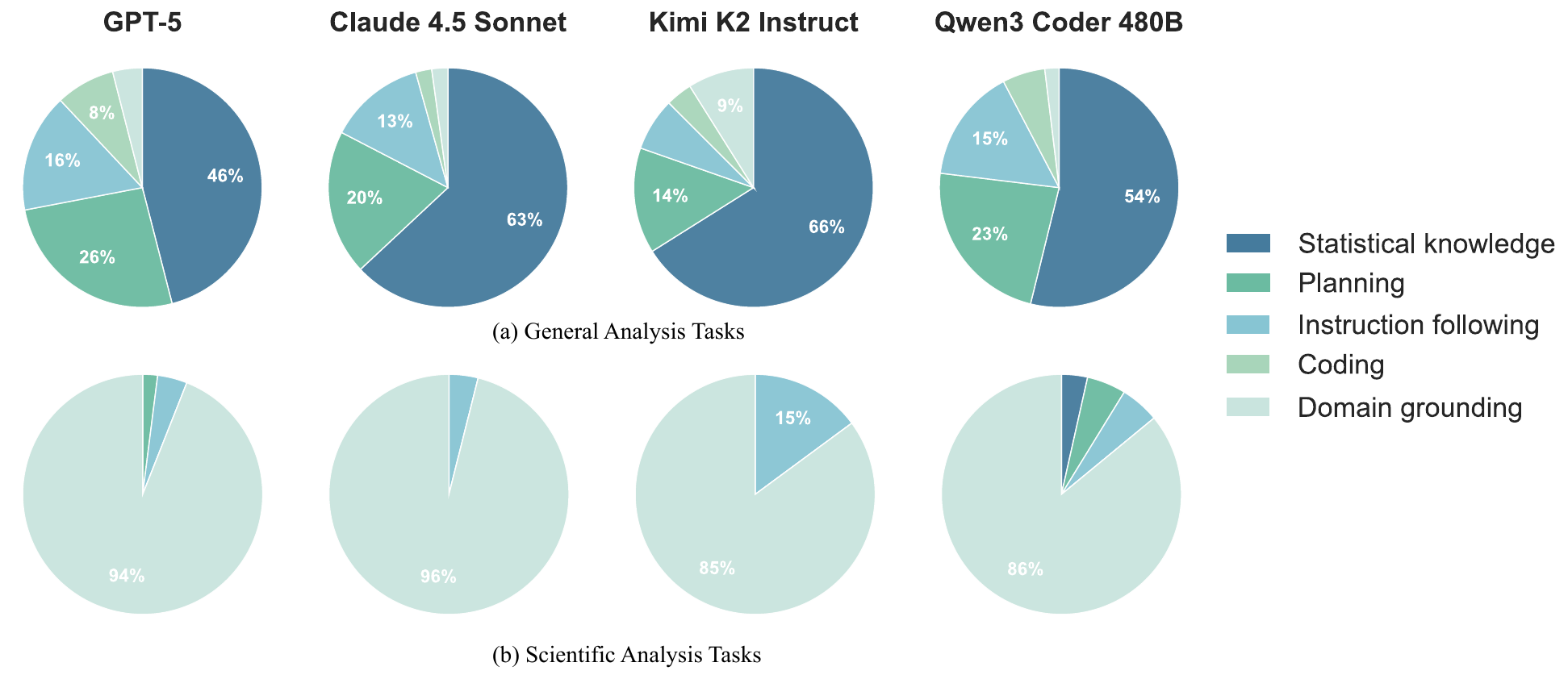}
    \caption{\textbf{Error type breakdowns for four LLMs} on (a) general analysis tasks (\textsc{QRData} and \textsc{DAEval}) and (b) scientific analysis tasks (\textsc{DSBio}).
    For each model and task family, we uniformly sample 50 failed trajectories and manually assign a primary error category (definitions in Appendix~\ref{sec:error_type}; representative cases in Appendix~\ref{app:failure case}).
    A key shift emerges: while failures on general tasks are dominated by statistical knowledge and planning issues, failures on \textsc{DSBio} are overwhelmingly driven by domain-grounding errors (85--96\% across models).
    }
    \label{fig:error}
\end{figure*}

\begin{figure*}[t]
    \centering
    \includegraphics[width=\textwidth]{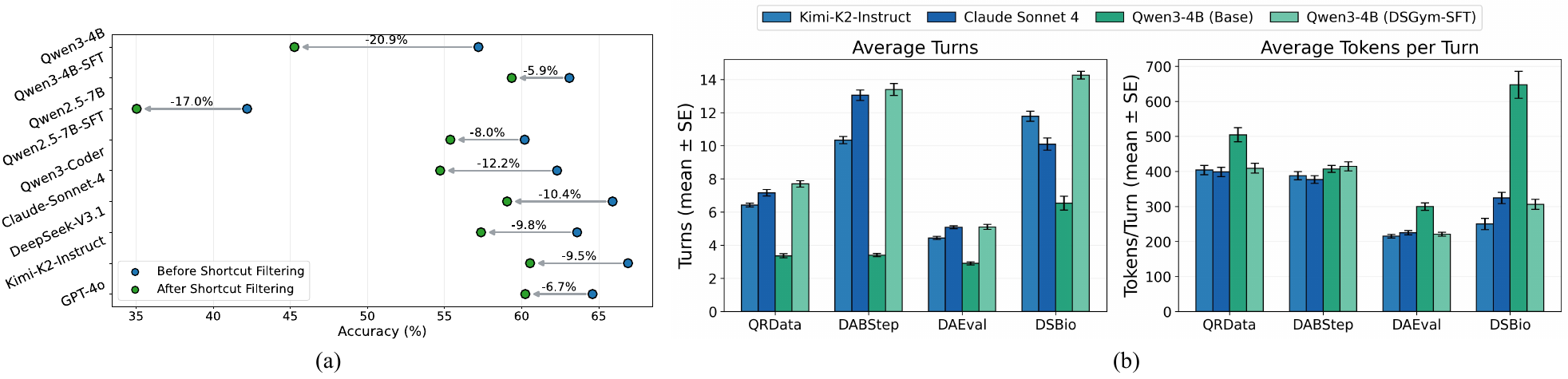}
    \caption{
        (a) \textbf{Accuracy on the same error-cleaned QRData split \emph{with vs. without} enforcing data dependency.} All models exhibit consistent drops after filtering, indicating that a non-trivial portion of pre-filter performance can be achieved via non-data-grounded shortcuts (e.g., memorization, priors, etc). (b) \textbf{Execution-grounded SFT changes agent interaction behavior toward teacher-like trajectories.} Across four datasets, we report the mean$\pm$std of the number of turns per trajectory and tokens per turn for two teacher models and a 4B base model before/after DSGym-SFT. DSGym-SFT increases the number of turns while shifting tokens-per-turn toward teacher-like statistics, indicating finer-grained decomposition and more iterative execution.
    }
    \label{fig:agent_behavior}
\end{figure*}

Table~\ref{tab:data-analysis} presents the accuracy on standardized benchmarks. Notably, \textsc{Kimi-K2-Instruct} and \textsc{Claude 4.5 Sonnet} perform relatively better than other models. We observe a universal performance drop on the \textsc{DABStep-Hard} compared to other easy splits, indicating that multi-step reasoning with heavy data dependencies remains a bottleneck even for frontier models.

On the expert-derived \textsc{DSBio} suite (Table~\ref{tab:scienceagent}), performance is consistently lower than on general tasks. Notably, \textsc{Kimi-K2-Instruct} achieves the best overall performance (43.33\%), followed by \textsc{Claude 4.5 Sonnet}, showcasing their relative robustness in utilizing specialized bioinformatics toolchains.

Finally, we assess end-to-end modeling capabilities in Table~\ref{tab:kaggle-analysis}. On \textsc{MLEBench-lite} and \textsc{DSPredict-Easy}, most frontier models achieve a near-perfect \textit{Valid Submission Rate} (>80\%), proving that they can reliably construct functional data pipelines. However, on \textbf{\textsc{DSPredict-Hard}}, even producing a valid submission remains a bottleneck, with most models failing to exceed 70\%. Furthermore, \textit{Medal Rates} across nearly all models are near zero, and the \textit{Median Rate} peaks at only 14.3\%. Among all evaluated models, \textbf{\textsc{GPT-5.1}} with high reasoning effort performs the best; we consistently observe that increasing reasoning effort for \textsc{GPT-5.1} leads to substantial gains across all prediction benchmarks.

\subsection{Analysis}

\textbf{Finding 1: A persistent scientific-domain gap remains even for frontier closed-source models.}

Despite strong performance on general-purpose data-analysis benchmarks (Table~\ref{tab:data-analysis}), all models substantially underperform on the \textsc{DSBio} suite (Table~\ref{tab:scienceagent}), which demands bioinformatics workflows and biologically grounded task interpretation (e.g., specialized libraries and modality-specific preprocessing). This gap suggests that frontier models still lack robust zero-shot grounding for realistic scientific analyses. 

Error breakdowns in Figure~\ref{fig:error} further indicate a qualitative shift in failure modes. More details including the error type definitions are deferred to Appendix~\ref{sec:error_type}. On general analysis tasks, failures are largely attributable to statistical-knowledge and planning issues; however, on \textsc{DSBio}, domain-grounding errors dominate across all models (85--96\% of sampled failures), with representative examples provided in Appendix~\ref{app:failure case}.

Our detailed analysis indicates that these biological grounding failures largely arise from two sources. First, agents often struggle to robustly interpret complex queries together with dataset metadata in the intended biological context. Since \textsc{DSBio} targets real-world, high-dimensional bioinformatics datasets from published studies, exploratory probing can surface unexpected signals that require specialized biological context; when this happens, agents frequently deviate from their initial plan and resort to trial-and-error reasoning with insufficient domain knowledge (see example in \ref{app:failure case}), ultimately producing incorrect answers. Second, agents exhibit limited familiarity with common bioinformatics methods and library usage. They may attempt to reimplement sophisticated algorithms from scratch rather than leveraging existing functions and libraries provided in the environment, and they often mishandle domain-specific edge cases intrinsic to biological data (e.g., sparsity), leading to missing steps or incorrect preprocessing and downstream analysis.

\textbf{Finding 2: Agents exhibit persistent behavioral limitations: Technical Constraints and Simplicity Bias.}

Beyond domain-specific knowledge gaps, our evaluation identifies technical constraints that hamper agent autonomy. These include \textbf{Environment Access Restrictions} (e.g., inability to install libraries or timeouts during large-scale training) and \textbf{API Incompatibilities}, manifested as version-specific errors such as hallucinating deprecated keyword arguments (e.g., \textit{early\_stopping\_rounds} in LightGBM).

However, these mechanical failures compound a more systemic issue: a simplicity bias. As shown in \cref{tab:kaggle-analysis}, agents exhibit a large delta between the valid rate (successful submission generation) and the above-median rate (outperforming humans). This gap is driven by \textbf{Low-Effort Heuristics}, where agents optimize for the path of least resistance—such as adopting a median-based baseline—rather than attempting rigorous, image-based modeling.

Ultimately, these three factors—environmental blocks, API friction, and internal heuristics—collectively drive the simplicity bias. When agents encounter technical resistance (environment or API errors), their preference for minimizing trajectory length leads them to abandon complex strategies in favor of superficial, safe analysis. This suggests that frontier models, while proficient at code generation, lack the "skeptical" persistence of expert data scientists, treating the first valid result as ground truth rather than a hypothesis to be improved. More details of our failure analysis can be found in \Cref{app:kaggle_error_type}.

\textbf{Finding 3: Shortcut filtering reveals substantial non-data-dependent solvability; smaller open-weight models are affected most.} As shown in Fig.~\ref{fig:agent_behavior}(a), enforcing data dependency consistently decreases accuracy across all evaluated models on the same error-cleaned QRData split (up to $\sim$21\% relative drop). 
Representative examples of tasks solvable without files are provided in Appendix~\ref{sec:shortcut_tasks}.

\section{Demonstration: Training Data Science Agents via \textsc{DSGym}}
\label{sec:training}

Beyond evaluation, \textsc{DSGym} also enables research on training data science agents with different algorithms such as supervised finetuning, curriculum learning, and reinforcement learning with the help of its distributed environment, standardized datasets and trajectory recording infrastructure. In this section, we demonstrate how \textsc{DSGym} can be used to construct high-quality synthetic training data through synthetic query construction and trajectory generation.
These procedures provide a practical example of leveraging the \textsc{DSGym} environment for agent training without human intervention.


\subsection{Execution-Grounded Data Synthesis}

We adopt a multi-stage process grounded in execution at every step to synthesize training data.

\textbf{Stage 1: Exploratory Query Generation.} 
To ensure generated questions are grounded in reality, we employ an ``Explore-and-Validate'' method. We utilize the default agent scaffold in \textsc{DSGym} for generating synthetic queries. The agent will be given an example query without ground-truth, context information, and dataset files, and then the agent can interact with the environment to come up with semantically distinct questions.
The agent is instructed to avoid trivial rephrasings and to design realistic tasks that can be solved through executable analysis. 
Critically, the agent is required to output not just the question, but also a reference Answer and strict answer format guidelines.
To fulfill this requirement, the generator agent must interact with the environment—loading data, inspecting schemas, and actually \emph{solving} its own proposed query via code execution. This self-validation step ensures that every synthesized query is feasible.

\textbf{Stage 2: Trajectory Sampling.}
Once the valid queries and their reference answers are obtained, we generate diverse solution paths. We instantiate a fresh \textsc{DSGym} environment for each query and use the default agent scaffold to generate $K$ independent candidate trajectories with temperature $T=0.8$.

\textbf{Stage 3: Joint Query-Trajectory Validation.}
We employ an LLM-based Judge to evaluate the {Query-Trajectory pair} as a coherent unit. 
Unlike simple answer matching, the judge evaluates the query and the whole trajectory using six execution-aware criteria:
\begin{itemize}
    \item \textbf{Query Clarity and Feasibility:} Is the query clearly-defined, unambiguous and realistically solvable?
    \item \textbf{Educational Value}: Does the query have learning value and sufficient complexity?
    \item \textbf{Exploratory Competence:} Does the trajectory perform sufficient data exploration?
    \item \textbf{Execution Robustness:} Are code blocks runnable? If errors occurred, did the agent successfully debug and recover?
    \item \textbf{Task Alignment:} Does the executed logic actually address the specific intent of the query?
    \item \textbf{Answer Plausibility:} Is the derived answer consistently supported by the final execution outputs and consistent with the reference answer?
\end{itemize}
After this quality filtering, we apply a lightweight Diversity Filter based on semantic similarity to discard synthesized queries that are trivial rephrasings of the original seed example.

\paragraph{Applicability to Existing Benchmarks.} 
While described above as a full synthesis pipeline, the Trajectory Sampling and Verification stages (Stages 2 \& 3) function as a modular subsystem. They can be applied directly to \emph{existing tasks} to distill high-quality, execution-verified reasoning traces for SFT.

\begin{table*}[!ht]
\centering\small
\resizebox{0.93\textwidth}{!}{
\begin{tabular}{lccccc}
\toprule
\textbf{Model} & \textbf{QRData-Verified} (\%) & \textbf{DABStep-easy} (\%) & \textbf{DABStep-hard} (\%) & \textbf{DAEval-Verified} (\%) & \textbf{\textsc{DSBio}} (\%)\\
\midrule
GPT-4o & 60.24 & 73.61 & 7.41 & 92.26 & 33.33\\
Claude Sonnet 4.5 & 61.35 & \textbf{83.33} & \textbf{37.04} & 91.71 & 42.22\\
Claude Sonnet 4 & 59.06 & 81.94 & 31.75 & 90.91 & 36.67\\
Qwen3-Coder 480B & 54.72 & 75.00 & 14.29 & 90.61 & 34.44\\
Kimi K2 Instruct & 63.68 & 77.78 & 28.84 & 92.82 & 43.33 \\
\midrule
Qwen2.5-7B-Instruct & 35.04 & 47.22 & 2.38 & 50.56 & 5.56 \\
Datamind-7B & 49.00 & 68.06 & 2.38 & 85.79 & 15.56\\
Qwen3-4B-Instruct & 45.27 & 58.33 & 2.9 & 64.47 & 6.67\\
Jupyter Agent Qwen3 4B & - & $70.80^*$ & $3.4^*$ & - & - \\
\cellcolor{rankThird}\textbf{Qwen3-4B-DSGym-SFT-2k} & \cellcolor{rankThird}\textbf{59.36} & \cellcolor{rankThird}\textbf{77.78} & \cellcolor{rankThird}\textbf{33.07} & \cellcolor{rankThird}\textbf{86.19} & \cellcolor{rankThird}\textbf{21.11} \\
\bottomrule
\end{tabular}}
\vspace{-2mm}
\caption{ Accuracy performance comparison across data analysis tasks. * means we directly report the numbers in the original report.
}
\label{tab:train-result}
\vspace{-6mm}
\end{table*}

\subsection{Case Study: The \textsc{DSGym-SFT} Dataset}
\label{subsec:synthesis_case}
To demonstrate the utility of this pipeline, we constructed a demonstration training corpus.
Starting from a seed subset of \textsc{QRData} and \textsc{DABStep}, we prompted agents to explore the datasets and generate 3,700 synthetic query candidates. 
These were re-executed to obtain full reasoning traces.
After applying our Joint Query-Trajectory Filtering, we curated \textbf{2,000 high-quality pairs}. 
This dataset, denoted as \textsc{DSGym-SFT}, represents a fully synthetic, execution-verified instruction tuning corpus. 

This example illustrates how \textsc{DSGym} transforms from a purely evaluative benchmark into a closed-loop training ecosystem, enabling scalable generation, assessment, and refinement of data-science agents through realistic, executable analytical tasks.

\subsection{Experiments}
Table~\ref{tab:train-result} shows that a 4B model fine-tuned on \textsc{DSGym-SFT} attains competitive performance relative to substantially larger baselines, illustrating the potential of execution-grounded synthesis for data-efficient improvement.

\textbf{Data-efficient gains on analysis tasks.} Fine-tuning on \textsc{DSGym-SFT} yields consistent gains over the Qwen3-4B base model across benchmarks, with particularly large improvements on \textsc{DABStep-hard}. Notably, although \textsc{DSGym-SFT} is constructed only on \emph{general} data analysis tasks, it also improves performance on the out-of-domain \textsc{DSBio} benchmark, suggesting that the planning, reasoning, or decomposition-oriented behaviors learned from general analysis can transfer to scientific workflows beyond the training distribution.


\textbf{More structured interaction behavior.} Beyond accuracy, DSGym-SFT also changes how agents interact with the environment: as shown in Fig.~\ref{fig:agent_behavior}(b), SFT increases depth of exploration, promotes finer-grained decomposition, and encourages iterative execution, which likely contributes to improved performance on complex workflows in \textsc{DABStep-hard} and \textsc{DSBio}.

\textbf{Less reliance on shortcut solvability.} Fig.~\ref{fig:agent_behavior}(a) indicates that smaller open-weight models experience the largest performance degradations when shortcut solutions are removed. In contrast, \textsc{DSGym-SFT} models exhibit substantially smaller drops, suggesting improved robustness to shortcut-based answering.
\section{Related Works}

\subsection{Benchmarks for Data Science}
Assessing LLMs' data science capabilities has been actively studied in recent years. Early research focused on relatively simple code-generation tasks; for instance, \cite{lai2023ds} investigated introductory-to-intermediate data analysis problems restricted to the use of seven commonly used Python libraries (e.g., Numpy \citep{harris2020array} and Pandas \citep{reback2020pandas}), and \cite{yin2023natural} explored problems of similar difficulty in interactive data science notebook settings. Although these benchmarks support fast and automated evaluation, their simplicity limits to capture multi-step and interactive agent behaviors. This limitation has motivated subsequent work to incorporate more realistic and challenging components, including iterative reasoning/planning, statistics/domain knowledge, repeated code execution, and debugging within an interactive environment \citep{huang-etal-2024-da, hu2024infiagentdabench, majumder2024discoverybench, zhang2025datascibench, lu2025stateval}. As a few representative examples, in data analysis tasks, \cite{liu-etal-2024-llms} curated reasoning tasks from statistics textbooks that require both data input/output processes and data exploration, \cite{yang-etal-2024-matplotagent} introduced a benchmark framework for evaluating LLMs' visualization ability, \cite{egg2025dabstep} examined financial data analysis involving multi-step reasoning over heterogeneous data sources, \cite{jing2024dsbenchfardatascience} studied agent behavior under long-context settings, and \cite{gu2024blade} considered open-ended data science questions collected from scientific literature. In predictive modeling tasks, \cite{chan2024mle-bench} curated 75 Kaggle competitions and examined how well LLM-based agents handle end-to-end ML engineering tasks.
\dsgym focuses more on providing a gym environment tailored to data science tasks, standardizing heterogeneous data and model interfaces.  

\subsection{Agents for Data Science}
Alongside benchmarks, a growing body of work studies agent scaffolds; how to structure agents to handle complex data science workflows \cite{rahman2025llm}. Many early approaches, including \cite{huang2023mlagentbench,hu2024infiagentdabench}, rely on a single linear execution trace as variations of ReAct or CodeAct \citep{yao2022react, wang2024executable} and have shown promising abilities of these agents. Recently, \cite{aide2025} improved upon this paradigm by representing candidate solutions as nodes in a tree. This tree representation enables the agent to explore multiple candidate solutions in parallel, backtrack from suboptimal trajectories, and refine the final solution. \cite{yang2025rdagentllmagentframeworkautonomous} further enhanced this scaffold with more sophisticated planning/reasoning modules, in which the agent generates ideas and verifies them multiple times before implementation. This approach has been shown to be effective for developing predictive models and achieves competitive performance on MLE-bench \citep{chan2024mle-bench}. Overall, the agent performance highly depends on how the system structures iteration and reasoning; effective agents explicitly conduct multi-step search over hypotheses, candidate solutions, and evaluation feedback.

Beyond scaffold-based approaches, many studies have explored the design of data science agents from multiple perspectives, including environment modeling, agent coordination, and task representation. \cite{you2025datawiseagent} developed an agent that can interact on a sequence of markdown or executable code cells in Jupyter Notebook environments. \cite{li2024autokaggle} considered a multi-agent system capable of completing end-to-end data science workflows, ranging from data preprocessing to report generation. From data-structural perspectives, \cite{hong2025data} considered representing a data science task as a graph, dynamically decomposing the main task into dependent subtasks and revising the graph as new evidence or constraints appear. The main goal of \textsc{DSGym} is to provide an easy-to-use, standardized system that supports reproducible training and evaluation of agent systems, making it simple for these agents to be adopted and assessed.
\section{Discussion and Limitations}

Our findings highlight both the opportunities and ongoing challenges in leveraging LLMs as agents for automated data science. We now discuss the  several avenues for improvement:
\begin{itemize}
    \item \textbf{Extending to RL.}
A key advantage of \textsc{DSGym} is its distributed, containerized, stateful execution, which naturally supports interactive optimization of agent policies. This makes \textsc{DSGym} a suitable environment for studying RL-style training and evaluation across multiple data science datasets. However, two challenges remain central: \emph{training signal design} and \emph{data and task coverage}. Existing data science trajectories are limited in scale, uneven in quality, and often underrepresent domain-specific scientific workflows. Moreover, providing informative credit assignment under sparse, long-horizon rewards remains an open problem. \textsc{DSGym} exposes these challenges in a controlled setting, enabling systematic investigation of reward design and verification-based filtering.



\item \textbf{Deepening Scientific Grounding.}
Our analysis on \textsc{DSBio} shows that generalist models struggle with domain-specific ontologies, data modalities, and tooling, with the gap particularly pronounced for smaller models. Two complementary directions may help address this limitation. Tool-oriented abstractions can reduce avoidable workflow errors by exposing robust domain primitives, while domain-adaptive learning (e.g., continued pretraining or finetuning on scientific corpora and verified analysis traces) may be necessary to improve conceptual grounding and method selection. Expanding to additional scientific domains (e.g., chemistry, materials science, or astronomy) is also important, not merely to increase task diversity, but to probe qualitatively different forms of domain knowledge under a standardized evaluation interface.

\item \textbf{Deterministic evaluation and open-ended discovery.}
We intentionally prioritize reproducibility through strict data dependency and deterministic evaluation metrics. However, many real-world scientific workflows are open-ended, involving stochastic outcomes, visualization, or multiple valid interpretations. \textsc{DSGym} currently does not cover such settings, including visualization-centric or exploratory tasks. Extending evaluation beyond deterministic regimes remains challenging and will likely require reliable validation mechanisms grounded in execution traces, such as carefully controlled LLM-based judges.

\item \textbf{\textsc{DSGym} as a live testbed.}
We envision \textsc{DSGym} as a living testbed that evolves with scientific tooling and emerging evaluation needs, complementing static benchmarks that are prone to memorization and rapid saturation. 
This live-but-auditable design supports reproducible measurement, systematic ablations, and principled tracking of progress over time.
\end{itemize}

\section{Conclusion}
We introduce \textsc{DSGym}, a standardized and extensible framework for evaluating data science agents in stateful, isolated execution environments. \textsc{DSGym} unifies heterogeneous benchmarks behind a single abstraction and supports reproducible end-to-end measurement of whether agents can plan, write, and execute scientific workflows while interacting with real data files. Crucially, \textsc{DSGym} revisits a core assumption in existing evaluations that file-grounded benchmarks necessarily measure data-dependent reasoning and provides tooling to mitigate prompt-only shortcut solvability. To support rigorous evaluation, we release \textsc{DSGym-Tasks}, which (i) standardizes and audits widely used analysis benchmarks and applies shortcut-solvability filtering, and (ii) expands coverage with domain-grounded scientific analysis tasks (\textsc{DSBio}) and realistic, challenging end-to-end modeling tasks (\textsc{DSPredict}). Through a systematic study of frontier proprietary and open-weight LLMs, we highlight persistent weaknesses in domain-specific workflows and recurring behaviors such as simplicity bias and insufficient verification. Finally, beyond evaluation, we show that \textsc{DSGym}'s execution environment can also be used to synthesize execution-verified trajectories for finetuning, illustrating a practical path toward improving data science agents. We hope \textsc{DSGym} serves as a live, auditable testbed that evolves with scientific practice while providing a moving yet reproducible target for evaluating and advancing LLM-based data science agents.



\section*{Acknowledgments}

We are extremly grateful to the Kaggle Team for providing us with access to resources for extended evaluations. We would also like to thank the members of Zou Group for helpful discussions.


\bibliography{reference}

@article{chan2024mle-bench,
  title={MLE-bench: Evaluating Machine Learning Agents on Machine Learning Engineering},
  author={Jun Shern Chan and Neil Chowdhury and Oliver Jaffe and James Aung and Dane Sherburn and Evan Mays and Giulio Starace and Kevin Liu and Leon Maksin and Tejal Patwardhan and Lilian Weng and Aleksander Mądry},
  year={2024},
  eprint={2410.07095},
  archivePrefix={arXiv},
  primaryClass={cs.CL},
  url={https://arxiv.org/abs/2410.07095}
}

@article{scientific,
	author = {Wang, Hanchen and Fu, Tianfan and Du, Yuanqi and Gao, Wenhao and Huang, Kexin and Liu, Ziming and Chandak, Payal and Liu, Shengchao and Van Katwyk, Peter and Deac, Andreea and Anandkumar, Anima and Bergen, Karianne and Gomes, Carla P. and Ho, Shirley and Kohli, Pushmeet and Lasenby, Joan and Leskovec, Jure and Liu, Tie-Yan and Manrai, Arjun and Marks, Debora and Ramsundar, Bharath and Song, Le and Sun, Jimeng and Tang, Jian and Veli{\v c}kovi{\'c}, Petar and Welling, Max and Zhang, Linfeng and Coley, Connor W. and Bengio, Yoshua and Zitnik, Marinka},
	da = {2023/08/01},
	doi = {10.1038/s41586-023-06221-2},
	id = {Wang2023},
	isbn = {1476-4687},
	journal = {Nature},
	number = {7972},
	pages = {47--60},
	title = {Scientific discovery in the age of artificial intelligence},
	ty = {JOUR},
	url = {https://doi.org/10.1038/s41586-023-06221-2},
	volume = {620},
	year = {2023}
}

@article{Wang_2024,
   title={A survey on large language model based autonomous agents},
   volume={18},
   ISSN={2095-2236},
   url={http://dx.doi.org/10.1007/s11704-024-40231-1},
   DOI={10.1007/s11704-024-40231-1},
   number={6},
   journal={Frontiers of Computer Science},
   publisher={Springer Science and Business Media LLC},
   author={Wang, Lei and Ma, Chen and Feng, Xueyang and Zhang, Zeyu and Yang, Hao and Zhang, Jingsen and Chen, Zhiyuan and Tang, Jiakai and Chen, Xu and Lin, Yankai and Zhao, Wayne Xin and Wei, Zhewei and Wen, Jirong},
   year={2024},
   month=mar }

@article{boiko2023emergent,
  title={Emergent autonomous scientific research capabilities of large language models},
  author={Boiko, Daniil A and MacKnight, Robert and Gomes, Gabe},
  journal={arXiv preprint arXiv:2304.05332},
  year={2023}
}

@inproceedings{lai2023ds,
  title={DS-1000: A natural and reliable benchmark for data science code generation},
  author={Lai, Yuhang and Li, Chengxi and Wang, Yiming and Zhang, Tianyi and Zhong, Ruiqi and Zettlemoyer, Luke and Yih, Wen-tau and Fried, Daniel and Wang, Sida and Yu, Tao},
  booktitle={International Conference on Machine Learning},
  pages={18319--18345},
  year={2023},
  organization={PMLR}
}

@misc{reback2020pandas,
    author       = {The pandas development team},
    title        = {pandas-dev/pandas: Pandas},
    month        = feb,
    year         = 2020,
    publisher    = {Zenodo},
    version      = {latest},
    doi          = {10.5281/zenodo.3509134},
    url          = {https://doi.org/10.5281/zenodo.3509134}
}

@article{         harris2020array,
 title         = {Array programming with {NumPy}},
 author        = {Charles R. Harris and K. Jarrod Millman and St{\'{e}}fan J.
                 van der Walt and Ralf Gommers and Pauli Virtanen and David
                 Cournapeau and Eric Wieser and Julian Taylor and Sebastian
                 Berg and Nathaniel J. Smith and Robert Kern and Matti Picus
                 and Stephan Hoyer and Marten H. van Kerkwijk and Matthew
                 Brett and Allan Haldane and Jaime Fern{\'{a}}ndez del
                 R{\'{i}}o and Mark Wiebe and Pearu Peterson and Pierre
                 G{\'{e}}rard-Marchant and Kevin Sheppard and Tyler Reddy and
                 Warren Weckesser and Hameer Abbasi and Christoph Gohlke and
                 Travis E. Oliphant},
 year          = {2020},
 month         = sep,
 journal       = {Nature},
 volume        = {585},
 number        = {7825},
 pages         = {357--362},
 doi           = {10.1038/s41586-020-2649-2},
 publisher     = {Springer Science and Business Media {LLC}},
 url           = {https://doi.org/10.1038/s41586-020-2649-2}
}

@misc{jing2024dsbenchfardatascience,
      title={DSBench: How Far Are Data Science Agents to Becoming Data Science Experts?}, 
      author={Liqiang Jing and Zhehui Huang and Xiaoyang Wang and Wenlin Yao and Wenhao Yu and Kaixin Ma and Hongming Zhang and Xinya Du and Dong Yu},
      year={2024},
      eprint={2409.07703},
      archivePrefix={arXiv},
      primaryClass={cs.AI},
      url={https://arxiv.org/abs/2409.07703}, 
}

@misc{hu2024infiagentdabench,
      title={InfiAgent-DABench: Evaluating Agents on Data Analysis Tasks}, 
      author={Xueyu Hu and Ziyu Zhao and Shuang Wei and Ziwei Chai and Guoyin Wang and Xuwu Wang and Jing Su and Jingjing Xu and Ming Zhu and Yao Cheng and Jianbo Yuan and Kun Kuang and Yang Yang and Hongxia Yang and Fei Wu},
      year={2024},
      eprint={2401.05507},
      archivePrefix={arXiv},
      primaryClass={cs.CL}
}

@inproceedings{huang-etal-2024-da,
    title = "{DA}-Code: Agent Data Science Code Generation Benchmark for Large Language Models",
    author = "Huang, Yiming  and
      Luo, Jianwen  and
      Yu, Yan  and
      Zhang, Yitong  and
      Lei, Fangyu  and
      Wei, Yifan  and
      He, Shizhu  and
      Huang, Lifu  and
      Liu, Xiao  and
      Zhao, Jun  and
      Liu, Kang",
    editor = "Al-Onaizan, Yaser  and
      Bansal, Mohit  and
      Chen, Yun-Nung",
    booktitle = "Proceedings of the 2024 Conference on Empirical Methods in Natural Language Processing",
    month = nov,
    year = "2024",
    address = "Miami, Florida, USA",
    publisher = "Association for Computational Linguistics",
    url = "https://aclanthology.org/2024.emnlp-main.748/",
    doi = "10.18653/v1/2024.emnlp-main.748",
    pages = "13487--13521"
}

@inproceedings{liu-etal-2024-llms,
    title = "Are {LLM}s Capable of Data-based Statistical and Causal Reasoning? Benchmarking Advanced Quantitative Reasoning with Data",
    author = "Liu, Xiao  and
      Wu, Zirui  and
      Wu, Xueqing  and
      Lu, Pan  and
      Chang, Kai-Wei  and
      Feng, Yansong",
    editor = "Ku, Lun-Wei  and
      Martins, Andre  and
      Srikumar, Vivek",
    booktitle = "Findings of the Association for Computational Linguistics ACL 2024",
    month = aug,
    year = "2024",
    address = "Bangkok, Thailand and virtual meeting",
    publisher = "Association for Computational Linguistics",
    url = "https://aclanthology.org/2024.findings-acl.548",
    pages = "9215--9235",
}

@article{lu2025stateval,
  title={StatEval: A Comprehensive Benchmark for Large Language Models in Statistics},
  author={Lu, Yuchen and Yang, Run and Zhang, Yichen and Yu, Shuguang and Dai, Runpeng and Wang, Ziwei and Xiang, Jiayi and Gao, Siran and Ruan, Xinyao and Huang, Yirui and others},
  journal={arXiv preprint arXiv:2510.09517},
  year={2025}
}

@article{egg2025dabstep,
  title={DABstep: Data Agent Benchmark for Multi-step Reasoning},
  author={Egg, Alex and Goyanes, Martin Iglesias and Kingma, Friso and Mora, Andreu and von Werra, Leandro and Wolf, Thomas},
  journal={arXiv preprint arXiv:2506.23719},
  year={2025}
}

@article{majumder2024discoverybench,
  title={Discoverybench: Towards data-driven discovery with large language models},
  author={Majumder, Bodhisattwa Prasad and Surana, Harshit and Agarwal, Dhruv and Mishra, Bhavana Dalvi and Meena, Abhijeetsingh and Prakhar, Aryan and Vora, Tirth and Khot, Tushar and Sabharwal, Ashish and Clark, Peter},
  journal={arXiv preprint arXiv:2407.01725},
  year={2024}
}

@article{zhang2025datascibench,
  title={Datascibench: An llm agent benchmark for data science},
  author={Zhang, Dan and Zhoubian, Sining and Cai, Min and Li, Fengzu and Yang, Lekang and Wang, Wei and Dong, Tianjiao and Hu, Ziniu and Tang, Jie and Yue, Yisong},
  journal={arXiv preprint arXiv:2502.13897},
  year={2025}
}

@article{huang2023mlagentbench,
  title={Mlagentbench: Evaluating language agents on machine learning experimentation},
  author={Huang, Qian and Vora, Jian and Liang, Percy and Leskovec, Jure},
  journal={arXiv preprint arXiv:2310.03302},
  year={2023}
}

@inproceedings{hong2025data,
  title={Data interpreter: An llm agent for data science},
  author={Hong, Sirui and Lin, Yizhang and Liu, Bang and Liu, Bangbang and Wu, Binhao and Zhang, Ceyao and Li, Danyang and Chen, Jiaqi and Zhang, Jiayi and Wang, Jinlin and others},
  booktitle={Findings of the Association for Computational Linguistics: ACL 2025},
  pages={19796--19821},
  year={2025}
}

@inproceedings{yin2023natural,
  title={Natural language to code generation in interactive data science notebooks},
  author={Yin, Pengcheng and Li, Wen-Ding and Xiao, Kefan and Rao, Abhishek and Wen, Yeming and Shi, Kensen and Howland, Joshua and Bailey, Paige and Catasta, Michele and Michalewski, Henryk and others},
  booktitle={Proceedings of the 61st Annual Meeting of the Association for Computational Linguistics (Volume 1: Long Papers)},
  pages={126--173},
  year={2023}
}

@article{aide2025,
      title={AIDE: AI-Driven Exploration in the Space of Code}, 
      author={Zhengyao Jiang and Dominik Schmidt and Dhruv Srikanth and Dixing Xu and Ian Kaplan and Deniss Jacenko and Yuxiang Wu},
      year={2025},
      eprint={2502.13138},
      archivePrefix={arXiv},
      primaryClass={cs.AI},
      url={https://arxiv.org/abs/2502.13138}, 
}

@inproceedings{wang2024executable,
      title={Executable Code Actions Elicit Better LLM Agents}, 
      author={Xingyao Wang and Yangyi Chen and Lifan Yuan and Yizhe Zhang and Yunzhu Li and Hao Peng and Heng Ji},
      year={2024},
      eprint={2402.01030},
      booktitle={ICML}
}

@misc{yang2025rdagentllmagentframeworkautonomous,
      title={R\&D-Agent: An LLM-Agent Framework Towards Autonomous Data Science}, 
      author={Xu Yang and Xiao Yang and Shikai Fang and Yifei Zhang and Jian Wang and Bowen Xian and Qizheng Li and Jingyuan Li and Minrui Xu and Yuante Li and Haoran Pan and Yuge Zhang and Weiqing Liu and Yelong Shen and Weizhu Chen and Jiang Bian},
      year={2025},
      eprint={2505.14738},
      archivePrefix={arXiv},
      primaryClass={cs.AI},
      url={https://arxiv.org/abs/2505.14738}, 
}

@inproceedings{llamafactory,
    title = "{L}lama{F}actory: Unified Efficient Fine-Tuning of 100+ Language Models",
    author = "Zheng, Yaowei  and
      Zhang, Richong  and
      Zhang, Junhao  and
      Ye, Yanhan  and
      Luo, Zheyan",
    editor = "Cao, Yixin  and
      Feng, Yang  and
      Xiong, Deyi",
    booktitle = "Proceedings of the 62nd Annual Meeting of the Association for Computational Linguistics (Volume 3: System Demonstrations)",
    month = aug,
    year = "2024",
    address = "Bangkok, Thailand",
    publisher = "Association for Computational Linguistics",
    url = "https://aclanthology.org/2024.acl-demos.38/",
    doi = "10.18653/v1/2024.acl-demos.38",
    pages = "400--410"
}

@misc{gpt-4o,
  title={Hello GPT-4o},
  note={\url{https://openai.com/index/hello-gpt-4o/}},
  author={OpenAI},
  year={2024}
}

@misc{gpt-5,
  title={Introducing GPT-5},
  note={\url{https://openai.com/index/introducing-gpt-5/}},
  author={OpenAI},
  year={2025}
}

@misc{v3.1,
  title={DeepSeek-V3.1 Release},
  note={\url{https://api-docs.deepseek.com/news/news250821}},
  author={DeepSeek-AI},
  year={2025}
}

@article{qwen3,
  author       = {An Yang and
                  Anfeng Li and
                  Baosong Yang and
                  Beichen Zhang and
                  Binyuan Hui and
                  Bo Zheng and
                  Bowen Yu and
                  Chang Gao and
                  Chengen Huang and
                  Chenxu Lv and
                  Chujie Zheng and
                  Dayiheng Liu and
                  Fan Zhou and
                  Fei Huang and
                  Feng Hu and
                  Hao Ge and
                  Haoran Wei and
                  Huan Lin and
                  Jialong Tang and
                  Jian Yang and
                  Jianhong Tu and
                  Jianwei Zhang and
                  Jian Yang and
                  et al.},
  title        = {Qwen3 Technical Report},
  journal      = {CoRR},
  volume       = {abs/2505.09388},
  year         = {2025},
  url          = {https://doi.org/10.48550/arXiv.2505.09388},
  doi          = {10.48550/ARXIV.2505.09388},
  eprinttype    = {arXiv},
  eprint       = {2505.09388},
  bibsource    = {dblp computer science bibliography, https://dblp.org}
}

@article{claude-4-5,
  title={Introducing Claude Sonnet 4.5},
  author={Anthropic},
  year={2025},
  note={Available at \url{https://www.anthropic.com/news/claude-sonnet-4-5}}
}

@article{claude-4,
  title={Introducing Claude 4},
  author={Anthropic},
  year={2025},
  note={Available at \url{https://www.anthropic.com/news/claude-4}}
}

@misc{kimiteam2025kimik2openagentic,
      title={Kimi K2: Open Agentic Intelligence}, 
      author={Kimi Team and Yifan Bai and Yiping Bao and Guanduo Chen and Jiahao Chen and Ningxin Chen and Ruijue Chen and Yanru Chen and Yuankun Chen and Yutian Chen and Zhuofu Chen and Jialei Cui and Hao Ding and Mengnan Dong and Angang Du and Chenzhuang Du and Dikang Du and Yulun Du and Yu Fan and Yichen Feng and Kelin Fu and Bofei Gao and Hongcheng Gao and Peizhong Gao and Tong Gao and Xinran Gu and Longyu Guan and Haiqing Guo and Jianhang Guo and Hao Hu and Xiaoru Hao and Tianhong He and Weiran He and Wenyang He and Chao Hong and Yangyang Hu and Zhenxing Hu and Weixiao Huang and Zhiqi Huang and Zihao Huang and Tao Jiang and Zhejun Jiang and Xinyi Jin and Yongsheng Kang and Guokun Lai and Cheng Li and Fang Li and Haoyang Li and Ming Li and Wentao Li and Yanhao Li and Yiwei Li and Zhaowei Li and Zheming Li and Hongzhan Lin and Xiaohan Lin and Zongyu Lin and Chengyin Liu and Chenyu Liu and Hongzhang Liu and Jingyuan Liu and Junqi Liu and Liang Liu and Shaowei Liu and T. Y. Liu and Tianwei Liu and Weizhou Liu and Yangyang Liu and Yibo Liu and Yiping Liu and Yue Liu and Zhengying Liu and Enzhe Lu and Lijun Lu and Shengling Ma and Xinyu Ma and Yingwei Ma and Shaoguang Mao and Jie Mei and Xin Men and Yibo Miao and Siyuan Pan and Yebo Peng and Ruoyu Qin and Bowen Qu and Zeyu Shang and Lidong Shi and Shengyuan Shi and Feifan Song and Jianlin Su and Zhengyuan Su and Xinjie Sun and Flood Sung and Heyi Tang and Jiawen Tao and Qifeng Teng and Chensi Wang and Dinglu Wang and Feng Wang and Haiming Wang and Jianzhou Wang and Jiaxing Wang and Jinhong Wang and Shengjie Wang and Shuyi Wang and Yao Wang and Yejie Wang and Yiqin Wang and Yuxin Wang and Yuzhi Wang and Zhaoji Wang and Zhengtao Wang and Zhexu Wang and Chu Wei and Qianqian Wei and Wenhao Wu and Xingzhe Wu and Yuxin Wu and Chenjun Xiao and Xiaotong Xie and Weimin Xiong and Boyu Xu and Jing Xu and Jinjing Xu and L. H. Xu and Lin Xu and Suting Xu and Weixin Xu and Xinran Xu and Yangchuan Xu and Ziyao Xu and Junjie Yan and Yuzi Yan and Xiaofei Yang and Ying Yang and Zhen Yang and Zhilin Yang and Zonghan Yang and Haotian Yao and Xingcheng Yao and Wenjie Ye and Zhuorui Ye and Bohong Yin and Longhui Yu and Enming Yuan and Hongbang Yuan and Mengjie Yuan and Haobing Zhan and Dehao Zhang and Hao Zhang and Wanlu Zhang and Xiaobin Zhang and Yangkun Zhang and Yizhi Zhang and Yongting Zhang and Yu Zhang and Yutao Zhang and Yutong Zhang and Zheng Zhang and Haotian Zhao and Yikai Zhao and Huabin Zheng and Shaojie Zheng and Jianren Zhou and Xinyu Zhou and Zaida Zhou and Zhen Zhu and Weiyu Zhuang and Xinxing Zu},
      year={2025},
      eprint={2507.20534},
      archivePrefix={arXiv},
      primaryClass={cs.LG},
      url={https://arxiv.org/abs/2507.20534}, 
}

@article{gpt-oss,
  title={Introducing gpt-oss},
  author={OpenAI},
  year={2025},
  note={Available at \url{https://openai.com/index/introducing-gpt-oss/}}
}

@misc{qwen2025qwen25technicalreport,
      title={Qwen2.5 Technical Report}, 
      author={Qwen and : and An Yang and Baosong Yang and Beichen Zhang and Binyuan Hui and Bo Zheng and Bowen Yu and Chengyuan Li and Dayiheng Liu and Fei Huang and Haoran Wei and Huan Lin and Jian Yang and Jianhong Tu and Jianwei Zhang and Jianxin Yang and Jiaxi Yang and Jingren Zhou and Junyang Lin and Kai Dang and Keming Lu and Keqin Bao and Kexin Yang and Le Yu and Mei Li and Mingfeng Xue and Pei Zhang and Qin Zhu and Rui Men and Runji Lin and Tianhao Li and Tianyi Tang and Tingyu Xia and Xingzhang Ren and Xuancheng Ren and Yang Fan and Yang Su and Yichang Zhang and Yu Wan and Yuqiong Liu and Zeyu Cui and Zhenru Zhang and Zihan Qiu},
      year={2025},
      eprint={2412.15115},
      archivePrefix={arXiv},
      primaryClass={cs.CL},
      url={https://arxiv.org/abs/2412.15115}, 
}

@article{qiao2025scaling,
  title={Scaling Generalist Data-Analytic Agents},
  author={Qiao, Shuofei and Zhao, Yanqiu and Qiu, Zhisong and Wang, Xiaobin and Zhang, Jintian and Bin, Zhao and Zhang, Ningyu and Jiang, Yong and Xie, Pengjun and Huang, Fei and others},
  journal={arXiv preprint arXiv:2509.25084},
  year={2025}
}

@Article{genes15030298,
AUTHOR = {Bhattacharya, Soumyaroop and Myers, Jacquelyn A. and Baker, Cameron and Guo, Minzhe and Danopoulos, Soula and Myers, Jason R. and Bandyopadhyay, Gautam and Romas, Stephen T. and Huyck, Heidie L. and Misra, Ravi S. and Dutra, Jennifer and Holden-Wiltse, Jeanne and McDavid, Andrew N. and Ashton, John M. and Al Alam, Denise and Potter, S. Steven and Whitsett, Jeffrey A. and Xu, Yan and Pryhuber, Gloria S. and Mariani, Thomas J.},
TITLE = {Single-Cell Transcriptomic Profiling Identifies Molecular Phenotypes of Newborn Human Lung Cells},
JOURNAL = {Genes},
VOLUME = {15},
YEAR = {2024},
NUMBER = {3},
ARTICLE-NUMBER = {298},
URL = {https://www.mdpi.com/2073-4425/15/3/298},
PubMedID = {38540357},
ISSN = {2073-4425},
DOI = {10.3390/genes15030298}
}

@Article{Zhao2025,
author={Zhao, Quanyi
and Pedroza, Albert
and Sharma, Disha
and Gu, Wenduo
and Dalal, Alex
and Weldy, Chad
and Jackson, William
and Li, Daniel Yuhang
and Ryan, Yana
and Nguyen, Trieu
and Shad, Rohan
and Palmisano, Brian T.
and Monteiro, Jo{\~a}o P.
and Worssam, Matthew
and Berezowitz, Alexa
and Iyer, Meghana
and Shi, Huitong
and Kundu, Ramendra
and Limbu, Lasemahang
and Kim, Juyong Brian
and Kundaje, Anshul
and Fischbein, Michael
and Wirka, Robert
and Quertermous, Thomas
and Cheng, Paul},
title={A cell and transcriptome atlas of human arterial vasculature},
journal={Cell Genomics},
year={2025},
month={Dec},
day={10},
publisher={Elsevier},
volume={5},
number={12},
issn={2666-979X},
doi={10.1016/j.xgen.2025.101034},
url={https://doi.org/10.1016/j.xgen.2025.101034}
}

@Article{Bhat-Nakshatri2024,
author={Bhat-Nakshatri, Poornima
and Gao, Hongyu
and Khatpe, Aditi S.
and Adebayo, Adedeji K.
and McGuire, Patrick C.
and Erdogan, Cihat
and Chen, Duojiao
and Jiang, Guanglong
and New, Felicia
and German, Rana
and Emmert, Lydia
and Sandusky, George
and Storniolo, Anna Maria
and Liu, Yunlong
and Nakshatri, Harikrishna},
title={Single-nucleus chromatin accessibility and transcriptomic map of breast tissues of women of diverse genetic ancestry},
journal={Nature Medicine},
year={2024},
month={Dec},
day={01},
volume={30},
number={12},
pages={3482-3494},
issn={1546-170X},
doi={10.1038/s41591-024-03011-9},
url={https://doi.org/10.1038/s41591-024-03011-9}
}

@Article{Alasoo2018,
author={Alasoo, Kaur
and Rodrigues, Julia
and Mukhopadhyay, Subhankar
and Knights, Andrew J.
and Mann, Alice L.
and Kundu, Kousik
and Hale, Christine
and Dougan, Gordon
and Gaffney, Daniel J.
and Consortium, H. I. P. S. C. I.},
title={Shared genetic effects on chromatin and gene expression indicate a role for enhancer priming in immune response},
journal={Nature Genetics},
year={2018},
month={Mar},
day={01},
volume={50},
number={3},
pages={424-431},
issn={1546-1718},
doi={10.1038/s41588-018-0046-7},
url={https://doi.org/10.1038/s41588-018-0046-7}
}

@Article{Schwartzentruber2018,
author={Schwartzentruber, Jeremy
and Foskolou, Stefanie
and Kilpinen, Helena
and Rodrigues, Julia
and Alasoo, Kaur
and Knights, Andrew J.
and Patel, Minal
and Goncalves, Angela
and Ferreira, Rita
and Benn, Caroline Louise
and Wilbrey, Anna
and Bictash, Magda
and Impey, Emma
and Cao, Lishuang
and Lainez, Sergio
and Loucif, Alexandre Julien
and Whiting, Paul John
and Gutteridge, Alex
and Gaffney, Daniel J.
and Consortium, H. I. P. S. C. I.},
title={Molecular and functional variation in iPSC-derived sensory neurons},
journal={Nature Genetics},
year={2018},
month={Jan},
day={01},
volume={50},
number={1},
pages={54-61},
issn={1546-1718},
doi={10.1038/s41588-017-0005-8},
url={https://doi.org/10.1038/s41588-017-0005-8}
}

@article{
doi:10.1126/sciimmunol.adn3954,
author = {Joshua I. Gray  and Daniel P. Caron  and Steven B. Wells  and Rebecca Guyer  and Peter Szabo  and Daniel Rainbow  and Can Ergen  and Ksenia Rybkina  and Marissa C. Bradley  and Rei Matsumoto  and Kalpana Pethe  and Masaru Kubota  and Sarah Teichmann  and Joanne Jones  and Nir Yosef  and Mark Atkinson  and Maigan Brusko  and Todd M. Brusko  and Thomas J. Connors  and Peter A. Sims  and Donna L. Farber },
title = {Human {$\gamma\delta$} T cells in diverse tissues exhibit site-specific maturation dynamics across the life span},
journal = {Science Immunology},
volume = {9},
number = {96},
pages = {eadn3954},
year = {2024},
doi = {10.1126/sciimmunol.adn3954},
URL = {https://www.science.org/doi/abs/10.1126/sciimmunol.adn3954},
eprint = {https://www.science.org/doi/pdf/10.1126/sciimmunol.adn3954}}

@article{10.1084/jem.20191130,
    author = {Wang, Yalong and Song, Wanlu and Wang, Jilian and Wang, Ting and Xiong, Xiaochen and Qi, Zhen and Fu, Wei and Yang, Xuerui and Chen, Ye-Guang},
    title = {Single-cell transcriptome analysis reveals differential nutrient absorption functions in human intestine},
    journal = {Journal of Experimental Medicine},
    volume = {217},
    number = {2},
    pages = {e20191130},
    year = {2019},
    month = {11},
    issn = {0022-1007},
    doi = {10.1084/jem.20191130},
    url = {https://doi.org/10.1084/jem.20191130},
    eprint = {https://rupress.org/jem/article-pdf/217/2/e20191130/1769254/jem_20191130.pdf},
}

@misc{chen2025largelanguagemodelbaseddata,
      title={Large Language Model-based Data Science Agent: A Survey}, 
      author={Ke Chen and Peiran Wang and Yaoning Yu and Xianyang Zhan and Haohan Wang},
      year={2025},
      eprint={2508.02744},
      archivePrefix={arXiv},
      primaryClass={cs.AI},
      url={https://arxiv.org/abs/2508.02744}, 
}

@article{Sun_2025,
   title={A Survey on Large Language Model-based Agents for Statistics and Data Science},
   ISSN={1537-2731},
   url={http://dx.doi.org/10.1080/00031305.2025.2561140},
   DOI={10.1080/00031305.2025.2561140},
   journal={The American Statistician},
   publisher={Informa UK Limited},
   author={Sun, Maojun and Han, Ruijian and Jiang, Binyan and Qi, Houduo and Sun, Defeng and Yuan, Yancheng and Huang, Jian},
   year={2025},
   month=oct, pages={1–14} }

@inproceedings{yao2022react,
  title={React: Synergizing reasoning and acting in language models},
  author={Yao, Shunyu and Zhao, Jeffrey and Yu, Dian and Du, Nan and Shafran, Izhak and Narasimhan, Karthik R and Cao, Yuan},
  booktitle={The eleventh international conference on learning representations},
  year={2022}
}

@inproceedings{yang-etal-2024-matplotagent,
    title = "{M}at{P}lot{A}gent: Method and Evaluation for {LLM}-Based Agentic Scientific Data Visualization",
    author = "Yang, Zhiyu  and
      Zhou, Zihan  and
      Wang, Shuo  and
      Cong, Xin  and
      Han, Xu  and
      Yan, Yukun  and
      Liu, Zhenghao  and
      Tan, Zhixing  and
      Liu, Pengyuan  and
      Yu, Dong  and
      Liu, Zhiyuan  and
      Shi, Xiaodong  and
      Sun, Maosong",
    editor = "Ku, Lun-Wei  and
      Martins, Andre  and
      Srikumar, Vivek",
    booktitle = "Findings of the Association for Computational Linguistics: ACL 2024",
    month = aug,
    year = "2024",
    address = "Bangkok, Thailand",
    publisher = "Association for Computational Linguistics",
    url = "https://aclanthology.org/2024.findings-acl.701/",
    doi = "10.18653/v1/2024.findings-acl.701",
    pages = "11789--11804"
}

@article{gu2024blade,
  title={Blade: Benchmarking language model agents for data-driven science},
  author={Gu, Ken and Shang, Ruoxi and Jiang, Ruien and Kuang, Keying and Lin, Richard-John and Lyu, Donghe and Mao, Yue and Pan, Youran and Wu, Teng and Yu, Jiaqian and others},
  journal={arXiv preprint arXiv:2408.09667},
  year={2024}
}

@inproceedings{you2025datawiseagent,
  title={DatawiseAgent: A Notebook-Centric LLM Agent Framework for Adaptive and Robust Data Science Automation},
  author={You, Ziming and Zhang, Yumiao and Xu, Dexuan and Lou, Yiwei and Yan, Yandong and Wang, Wei and Zhang, Huamin and Huang, Yu},
  booktitle={Proceedings of the 2025 Conference on Empirical Methods in Natural Language Processing},
  pages={1099--1123},
  year={2025}
}

@article{rahman2025llm,
  title={Llm-based data science agents: A survey of capabilities, challenges, and future directions},
  author={Rahman, Mizanur and Bhuiyan, Amran and Islam, Mohammed Saidul and Laskar, Md Tahmid Rahman and Mahbub, Ridwan and Masry, Ahmed and Joty, Shafiq and Hoque, Enamul},
  journal={arXiv preprint arXiv:2510.04023},
  year={2025}
}

@article{li2024autokaggle,
  title={Autokaggle: A multi-agent framework for autonomous data science competitions},
  author={Li, Ziming and Zang, Qianbo and Ma, David and Guo, Jiawei and Zheng, Tuney and Liu, Minghao and Niu, Xinyao and Wang, Yue and Yang, Jian and Liu, Jiaheng and others},
  journal={arXiv preprint arXiv:2410.20424},
  year={2024}
}
\bibliographystyle{plain}


\newpage

\appendix
\onecolumn

\etocdepthtag.toc{mtappendix}

\setcounter{figure}{0}
\renewcommand\thefigure{S\arabic{figure}}
\setcounter{table}{0}
\renewcommand\thetable{S\arabic{table}}

\renewcommand{\contentsname}{Appendix}

\etocdepthtag.toc{mtappendix}
\etocsettagdepth{mtchapter}{none}
\etocsettagdepth{mtappendix}{subsection}

{\hypersetup{linkcolor=darkblue}
\tableofcontents
}

\clearpage




\section{Additional Details of \textsc{DSGym-Tasks}}

\subsection{Examples of Refinement of Existing Benchmarks}
\label{sec:refinement}

Here we provide examples of the tasks that we filter.

\begin{datasetexample}
\textbf{Dataset:} QRData 

\textbf{Task:} Which cause-and-effect relationship is more likely? Please answer with A, B, or C. 
\begin{quote}
A. L tibia pain causes L tibia pain \\
B. L tibia pain causes L tibia pain \\
C. No causal relationship exists
\end{quote}

\textbf{Provided Answer:} C

\textbf{Issue Identified:}  
The answer choices contain duplicated options (A and B are identical), making the task ill-defined.

\textbf{Action Taken:}  
This task is filtered out during dataset refinement due to invalid answer options.
\end{datasetexample}

\begin{datasetexample}
\textbf{Dataset:} DAEval

\textbf{Task:} Is there a significant difference in the total number of vaccinations administered per hundred people between countries that use different vaccines? \\

Constraints:

Only consider countries using Pfizer/BioNTech, Moderna, Oxford/AstraZeneca, and Johnson\&Johnson/Janssen. 

The country must have data without null values in the column of total vaccinations per hundred people.

Use One-Way Analysis of Variance (ANOVA) to test if there's significant difference among different vaccine groups. 

Consider the differences among vaccine groups to be significant if the p-value is less than 0.05. \\

Answer Format: \{

@significance\_of\_difference[significance]

@p\_value[p\_value]

Where `significance' is a string that can either be `yes' or `no' based on the conditions specified in the constraints.

Where `p\_value' is a number between 0 and 1, rounded to four decimal places.

\}\\

\textbf{Expected Output Format:}
\begin{quote}
\texttt{@significance\_of\_difference[significance]} \\
\texttt{@p\_value[p\_value]} \\
\end{quote}

\textbf{Provided Answer:} \texttt{[['significance\_of\_difference', 'no']]} \\

\textbf{Issue Identified:}  
The required \texttt{p\_value} field is missing from the provided answer, despite being explicitly required by the task format. 

\textbf{Action Taken:}  
This task is filtered out during dataset refinement due to incomplete ground-truth annotation.
\end{datasetexample}

Here we provide examples of the tasks that we refine.

\subsection{More details about \textsc{DSBio}}
\label{appendix:dsbio}
DSGym-bio is a curated benchmark of \textbf{90} data-science questions grounded in publicly available biomedical research datasets. Table~\ref{tab:dsgym_distribution} summarizes the domain distribution. The benchmark primarily focuses on \textbf{single-cell biology} (56/90), reflecting both its prominence in modern bioinformatics and the availability of many high-quality, reasonably sized public datasets that fit our agent environment. We additionally include problems from \textbf{genetics} (21/90) and \textbf{spatial transcriptomics} (13/90) to broaden coverage across biomedical modalities. Table~\ref{tab:paper_list} lists representative research papers used to construct DSGym-bio, along with the number of problems derived from each paper, their domain labels, and the corresponding data sources. 

\begin{table}[h]
\centering
\caption{Distribution of question domains in the DSGym-bio dataset.}
\label{tab:dsgym_distribution}
\begin{tabular}{lrc}
\toprule
\textbf{Category} & \textbf{Count}  \\
\midrule
Single-cell biology      & 56 \\
Genetics                 & 21 \\
Spatial transcriptomics  & 13 \\
\midrule
\textbf{Total}           & \textbf{90} \\
\bottomrule
\end{tabular}
\end{table}

\begin{table}[t]
    \centering
    \caption{Overview of research papers included in \textbf{DSGym-bio}, with their publication venues.}
    \label{tab:paper_list}
    \rowcolors{2}{gray!5}{white}
    \renewcommand{\arraystretch}{1.1}
    \setlength{\tabcolsep}{4pt}
    \begin{small}
    \begin{tabularx}{\textwidth}{@{}
        >{\RaggedRight\itshape\arraybackslash}X
        >{\centering\arraybackslash}m{2.2cm}
        >{\centering\arraybackslash}m{2.2cm}
        >{\centering\arraybackslash}m{2.8cm}
    @{}}
        \toprule
        \textbf{Paper Title} & \textbf{Problem Count} & \textbf{Domain} & \textbf{Data Source} \\
        \midrule
        Single-Cell Transcriptomic Profiling Identifies Molecular Phenotypes of Newborn Human Lung Cells~\citep{genes15030298}  & 13 & Single-cell biology & \href{https://cellxgene.cziscience.com/collections/28e9d721-6816-48a2-8d0b-43bf0b0c0ebc}{cellxgene} \\
        A cell and transcriptome atlas of human arterial vasculature~\citep{Zhao2025} & 13 & Spatial Transcriptomics & \href{https://cellxgene.cziscience.com/collections/8f17ac63-aaba-44b5-9b78-60f121da4c2f}{cellxgene}\\
        Single-nucleus chromatin accessibility and transcriptomic map of breast tissues of women of diverse genetic ancestry~\citep{Bhat-Nakshatri2024} & 21 & Single-cell biology & \href{https://cellxgene.cziscience.com/collections/77446b76-1c2d-4a71-8e59-0efd4374d98e}{cellxgene}\\
        Shared genetic effects on chromatin and gene expression indicate a role for enhancer priming in immune response~\citep{Alasoo2018} & 9 & Genetics & \href{https://zenodo.org/records/259661}{zenodo}\\
        Molecular and functional variation in iPSC-derived sensory neurons~\citep{Schwartzentruber2018} & 12 & Genetics & \href{https://www.ebi.ac.uk/biostudies/studies/S-BSST16}{EMBL-EBI} \\
        Human $\gamma\delta$ T cells in diverse tissues exhibit site-specific maturation dynamics across the life span~\citep{doi:10.1126/sciimmunol.adn3954} & 12 & Single-cell biology & \href{https://cellxgene.cziscience.com/collections/ec691f5f-0aac-433c-8f78-e7f4b85a05e0}{cellxgene} \\
        Single-cell transcriptome analysis reveals differential nutrient absorption functions in human intestine~\citep{10.1084/jem.20191130} & 10 & Single-cell biology & \href{https://cellxgene.cziscience.com/collections/ff668d5d-5b3f-49ee-a007-ff0664bf35ec}{cellxgene}
        \\
        \bottomrule
    \end{tabularx}
    \end{small}
\end{table}

\subsection{Examples of \textsc{DSBio}}
\begin{datasetexample}
\textbf{Dataset:} DSGym-bio \\
\textbf{Domain:} Single-cell biology \\
\textbf{Task:} Identify co-expression modules in endothelial cells using hierarchical clustering on gene-gene correlation matrix. Using Pearson correlation on the top 500 most variable genes, cut the dendrogram at height 0.7 to define modules. How many genes belong to the largest co-expression module?\\
\textbf{Answer Guideline:} Answer must be a single numeric value (e.g., 42) with no units or text.\\
\textbf{Ground Truth Answer:} 19
\end{datasetexample}
\begin{datasetexample}
\textbf{Dataset:} DSGym-bio \\
\textbf{Domain:} Genetics \\
\textbf{Task:} Among response eQTL-caQTL pairs, what fraction shows chromatin QTL activity in naive macrophages before stimulation (fold change > 1.5)? Choose among these options: 20\%, 40\%, 60\%, or 80\%.\\
\textbf{Answer Guideline:} Answer must be one of the provided options exactly as shown, case-sensitive. For example: '20\%'.\\
\textbf{Ground Truth Answer:} 60\%
\end{datasetexample}
\begin{datasetexample}
\textbf{Dataset:} DSGym-bio \\
\textbf{Domain:} Spatial Transcriptomics \\
\textbf{Task:} Which scRNA-seq cell type shows the highest expression correlation with the largest spatial cluster in the aortic Slide-seqV2 dataset? Hint: look at 'author\_cell\_type' annotation in metadata.\\
\textbf{Answer Guideline:} Answer must be the exact single cell type name as shown in the metadata (e.g., 'Endothelial'), case-sensitive.\\
\textbf{Ground Truth Answer:} Smooth Muscle
\end{datasetexample}

\subsection{More details of Data Prediction Tasks}
We provide the full list of competitions in Table~\ref{tab:competition-list}.

\subsection{Details of Rule-Based Filtering for \textsc{DSPredict}}
\label{appendix:kaggle:filter}

A more detailed version of our rule-based filtering of Kaggle competitions is shown here. 
\begin{itemize}
\item Submissions must use CSV format to standardize automated submission handling and evaluation.
\item The competition must be a valid machine learning challenge (excluding CTFs and code golf tasks) to ensure relevance to data science modeling rather than puzzle solving or code optimization.
\item The dataset size must be under 15 GB to ensure feasible data loading and model training on typical research hardware.
\item The competition must have an available leaderboard to enable benchmarking and quantitative comparison of model performance.
\item The competition should require meaningful ML or data science engineering effort to solve, ensuring that it tests practical modeling, feature engineering, and pipeline design skills.
\item The competition description should be well-specified and solvable, providing clear objectives, evaluation criteria, and data structure to support reproducible experimentation.
\item Most of the competitions should not overlap with MLE Bench Lite.
\end{itemize}

\begin{longtable}{lll}
\caption{Competition Dataset Sizes categorized by difficulty and source} \\
\toprule
\textbf{Competition} & \textbf{Data Size} & \textbf{Domain} \\
\midrule
\endhead

\bottomrule
\endfoot

\midrule
\multicolumn{3}{c}{\textbf{\textsc{DSPredict-Easy}}} \\
\midrule
house-prices-advanced-regression-techniques & 956K & machine\_learning \\
playground-series-s3e1 & 6.2M & machine\_learning \\
playground-series-s3e11 & 48M & machine\_learning \\
playground-series-s3e13 & 292K & machine\_learning \\
playground-series-s3e14 & 2.9M & machine\_learning \\
playground-series-s3e15 & 1.7M & machine\_learning \\
playground-series-s3e16 & 8.7M & machine\_learning \\
playground-series-s3e19 & 13M & time\_series \\
playground-series-s3e21 & 672K & machine\_learning \\
playground-series-s3e22 & 388K & machine\_learning \\
playground-series-s3e24 & 22M & machine\_learning \\
playground-series-s3e25 & 2.1M & machine\_learning \\
playground-series-s3e26 & 1.4M & machine\_learning \\
playground-series-s3e3 & 456K & machine\_learning \\
playground-series-s3e5 & 232K & machine\_learning \\
playground-series-s3e7 & 3.7M & machine\_learning \\
playground-series-s3e9 & 488K & machine\_learning \\
playground-series-s4e1 & 21M & machine\_learning \\
playground-series-s4e10 & 6.0M & machine\_learning \\
playground-series-s4e11 & 27M & machine\_learning \\
playground-series-s4e12 & 318M & machine\_learning \\
playground-series-s4e2 & 4.4M & machine\_learning \\
playground-series-s4e3 & 5.3M & machine\_learning \\
playground-series-s4e4 & 8.1M & machine\_learning \\
playground-series-s4e5 & 43M & machine\_learning \\
playground-series-s4e6 & 16M & machine\_learning \\
playground-series-s4e7 & 1.1G & machine\_learning \\
playground-series-s4e8 & 285M & machine\_learning \\
playground-series-s4e9 & 46M & machine\_learning \\
playground-series-s5e1 & 21M & time\_series \\
playground-series-s5e2 & 39M & machine\_learning \\
playground-series-s5e3 & 188K & machine\_learning \\
playground-series-s5e4 & 91M & machine\_learning \\
playground-series-s5e5 & 48M & machine\_learning \\
playground-series-s5e6 & 49M & machine\_learning \\
playground-series-s5e7 & 1.1M & machine\_learning \\
playground-series-s5e8 & 86M & machine\_learning \\
titanic & 100K & machine\_learning \\

\midrule
\multicolumn{3}{c}{\textbf{\textsc{DSPredict-Hard}}} \\
\midrule
ashrae-energy-prediction & 2.5G & time\_series \\
career-con-2019 & 95M & sensor\_signal \\
champs-scalar-coupling & 1.6G & chemistry \\
data-science-bowl-2018 & 480M & computer\_vision \\
digit-recognizer & 123M & computer\_vision \\
elo-merchant-category-recommendation & 2.9G & business \\
gendered-pronoun-resolution & 7.5M & nlp \\
geolifeclef-2024 & 3.3G & geology \\
google-smartphone-decimeter-challenge & 12G & sensor\_signal \\
home-credit-default-risk & 2.5G & machine\_learning \\
home-data-for-ml-course & 1.2M & machine\_learning \\
humpback-whale-identification & 5.7G & computer\_vision \\
ieee-fraud-detection & 1.3G & machine\_learning \\
imaterialist-challenge-fashion-2018 & 378M & computer\_vision \\
imaterialist-challenge-furniture-2018 & 47M & computer\_vision \\
inclusive-images-challenge & 16G & computer\_vision \\
LANL-Earthquake-Prediction & 9.8G & sensor\_signal \\
liverpool-ion-switching & 140M & biology \\
m5-forecasting-accuracy & 430M & time\_series \\
m5-forecasting-uncertainty & 492M & time\_series \\
march-machine-learning-mania-2023 & 138M & sports \\
march-machine-learning-mania-2025 & 175M & sports \\
mens-machine-learning-competition-2018 & 1.6G & sports \\
mens-machine-learning-competition-2019 & 1.8G & sports \\
mens-march-mania-2022 & 228M & sports \\
microsoft-malware-prediction & 7.9G & machine\_learning \\
nlp-getting-started & 1.4M & nlp \\
novozymes-enzyme-stability-prediction & 16M & chemistry \\
open-problems-single-cell-perturbations & 4.3G & bioinformatics \\
otto-recommender-system & 12G & recommender\_system \\
pku-autonomous-driving & 5.9G & computer\_vision \\
planttraits2024 & 3.4G & computer\_vision \\
predict-ai-model-runtime & 6.9G & machine\_learning \\
recruit-restaurant-visitor-forecasting & 136M & time\_series \\
rsna-pneumonia-detection-challenge & 3.8G & computer\_vision \\
santander-customer-transaction-prediction & 579M & machine\_learning \\
santander-value-prediction-challenge & 1.1G & machine\_learning \\
siim-acr-pneumothorax-segmentation & 426M & computer\_vision \\
spaceship-titanic & 1.2M & machine\_learning \\
sp-society-camera-model-identification & 11G & computer\_vision \\
stanford-covid-vaccine & 2.6G & bioinformatics \\
statoil-iceberg-classifier-challenge & 1.7G & computer\_vision \\
store-sales-time-series-forecasting & 120M & time\_series \\
talkingdata-adtracking-fraud-detection & 11G & machine\_learning \\
tensorflow-speech-recognition-challenge & 6.9G & audio\_speech \\
tgs-salt-identification-challenge & 720M & computer\_vision \\
trec-covid-information-retrieval & 13G & nlp \\
understanding\_cloud\_organization & 6.0G & computer\_vision \\
ventilator-pressure-prediction & 667M & sensor\_signal \\
vsb-power-line-fault-detection & 12G & sensor\_signal \\
web-traffic-time-series-forecasting & 2.3G & time\_series \\
womens-machine-learning-competition-2019 & 19M & sports \\
youtube8m-2018 & 1.1G & computer\_vision \\
youtube8m-2019 & 534M & computer\_vision \\

\midrule
\multicolumn{3}{c}{\textbf{MLEBench-Lite}} \\
\midrule
aerial-cactus-identification & 236M & computer\_vision \\
aptos2019-blindness-detection & 18G & computer\_vision \\
denoising-dirty-documents & 239M & computer\_vision \\
detecting-insults-in-social-commentary & 4.3M & nlp \\
dog-breed-identification & 1.2G & computer\_vision \\
dogs-vs-cats-redux-kernels-edition & 2.0G & computer\_vision \\
histopathologic-cancer-detection & 13G & computer\_vision \\
jigsaw-toxic-comment-classification-challenge & 186M & nlp \\
leaf-classification & 64M & computer\_vision \\
mlsp-2013-birds & 1.2G & audio\_speech \\
new-york-city-taxi-fare-prediction & 6.9G & machine\_learning \\
nomad2018-predict-transparent-conductors & 21M & chemistry \\
plant-pathology-2020-fgvc7 & 1.2G & computer\_vision \\
random-acts-of-pizza & 17M & nlp \\
ranzcr-clip-catheter-line-classification & 19G & computer\_vision \\
siim-isic-melanoma-classification & 189G & computer\_vision \\
spooky-author-identification & 5.1M & nlp \\
tabular-playground-series-dec-2021 & 704M & machine\_learning \\
tabular-playground-series-may-2022 & 597M & machine\_learning \\
text-normalization-challenge-english-language & 745M & nlp \\
text-normalization-challenge-russian-language & 1.1G & nlp \\
the-icml-2013-whale-challenge-right-whale-redux & 1.6G & computer\_vision \\
\label{tab:competition-list}
\end{longtable}

\subsection{Examples of \textsc{DSPredict}}


\begin{datasetexample}
\textbf{Dataset:} \textsc{DSPredict-Hard} \\
\textbf{Domain:}  \\
\textbf{Competition:} web-traffic-time-series-forecasting
\begin{lstlisting}
**CHALLENGE NAME: web-traffic-time-series-forecasting**

Challenge description:
# Web Traffic Time Series Forecasting

## Competition Objective
Forecast future traffic to Wikipedia pages. This competition focuses on the problem of forecasting the future values of multiple time series, as it has always been one of the most challenging problems in the field. More specifically, we aim the competition at testing state-of-the-art methods designed by the participants, on the problem of forecasting future web traffic for approximately 145,000 Wikipedia articles.

Sequential or temporal observations emerge in many key real-world problems, ranging from biological data, financial markets, weather forecasting, to audio and video processing. The field of time series encapsulates many different problems, ranging from analysis and inference to classification and forecast.

This competition will run as two stages and involves prediction of actual future events. There will be a training stage during which the leaderboard is based on historical data, followed by a stage where participants are scored on real future events.

You have complete freedom in how to produce your forecasts: e.g. use of univariate vs multi-variate models, use of metadata (article identifier), hierarchical time series modeling (for different types of traffic), data augmentation (e.g. using Google Trends data to extend the dataset), anomaly and outlier detection and cleaning, different strategies for missing value imputation, and many more types of approaches.

We thank Google Inc. and Voleon for sponsorship of this competition, and Oren Anava and Vitaly Kuznetsov for organizing it.

Kaggle is excited to partner with research groups to push forward the frontier of machine learning. Research competitions make use of Kaggle's platform and experience, but are largely organized by the research group's data science team. Any questions or concerns regarding the competition data, quality, or topic will be addressed by them.

## Evaluation Criteria
Submissions are evaluated on SMAPE between forecasts and actual values. We define SMAPE = 0 when the actual and predicted values are both 0.

## Submission Requirements
For each article and day combination (see key.csv), you must predict the web traffic. The file should contain a header and have the following format:

Id,Visits
bf4edcf969af,0
929ed2bf52b9,0
ff29d0f51d5c0,etc.

Due to the large file size and number of rows, submissions may take a few minutes to score. Thank you for your patience.

## Prizes
$12,000
$8,000
$5,000

Top submissions will also have the opportunity to present their work at the NIPS Time Series Workshop in Long Beach, California, co-located with the top machine learning conference NIPS 2017. Attending the workshop is not required to participate in the competition, however only teams that are attending the workshop will be considered to present their work.

Attendees presenting in person are responsible for all costs associated with travel, expenses, and fees to attend NIPS 2017.

## Timeline
This competition has a training phase and a future forecasting phase. During the training phase, participants build models and predict on historical values. During the future phase, participants will forecast future traffic values.

September 1st, 2017 - Deadline to accept competition rules.
September 1st, 2017 - Team Merger deadline. This is the last day participants may join or merge teams.
September 1st, 2017 - Final dataset is released.
September 12th 7:59 PM UTC - Final submission deadline.

Competition winners will be revealed after November 13, 2017.

All deadlines are at 11:59 PM UTC on the corresponding day unless otherwise noted. The competition organizers reserve the right to update the contest timeline if they deem it necessary.

## Competition Details
- **Competition Host**: Google
- **Competition Type**: Research Prediction Competition
- **Start Date**: July 13, 2017
- **Close Date**: November 15, 2017
- **Total Prize Pool**: $25,000

## Citation
Maggie, Oren Anava, Vitaly Kuznetsov, and Will Cukierski. Web Traffic Time Series Forecasting. https://kaggle.com/competitions/web-traffic-time-series-forecasting, 2017. Kaggle.
\end{lstlisting}
\textbf{Dataset Description:}
\begin{lstlisting}
Web Traffic Time Series Forecasting Forecast future traffic to Wikipedia pages Dataset Description The training dataset consists of approximately 145k time series. Each of these time series represent a number of daily views of a different Wikipedia article, starting from July, 1st, 2015 up until December 31st, 2016. The leaderboard during the training stage is based on traffic from January, 1st, 2017 up until March 1st, 2017. The second stage will use training data up until September 1st, 2017. The final ranking of the competition will be based on predictions of daily views between September 13th, 2017 and November 13th, 2017 for each article in the dataset. You will submit your forecasts for these dates by September 12th. For each time series, you are provided the name of the article as well as the type of traffic that this time series represent (all, mobile, desktop, spider). You may use this metadata and any other publicly available data to make predictions. Unfortunately, the data source for this dataset does not distinguish between traffic values of zero and missing values. A missing value may mean the traffic was zero or that the data is not available for that day. To reduce the submission file size, each page and date combination has been given a shorter Id. The mapping between page names and the submission Id is given in the key files. File descriptions Files used for the first stage will end in '_1'. Files used for the second stage will end in '_2'. Both will have identical formats. The complete training data for the second stage will be made available prior to the second stage. train_*.csv- contains traffic data. This a csv file where each row corresponds to a particular article and each column correspond to a particular date. Some entries are missing data. The page names contain the Wikipedia project (e.g. en.wikipedia.org), type of access (e.g. desktop) and type of agent (e.g. spider). In other words, each article name has the following format: 'name_project_access_agent' (e.g. 'AKB48_zh.wikipedia.org_all-access_spider'). key_*.csv- gives the mapping between the page names and the shortened Id column used for prediction sample_submission_*.csv- a submission file showing the correct format License Subject to Competition Rules
\end{lstlisting}
\end{datasetexample}

\section{Case Studies}

\subsection{Case Studies of Data Analysis Tasks}\label{app:failure case}








\begin{failurecase}{Qwen3-Coder-480B}{DSGym-Bio (Single-cell biology)}
\vspace{-1mm}

\noindent\textbf{Keywords.} Jensen-Shannon divergence, gene expression normalization, highly variable gene identification.

\noindent\textbf{Task.} Calculate the Jensen-Shannon distance between gene expression distributions of cells from 'blood' vs 'lung' tissues. Use only the top 1000 most variable genes and normalize expression values to probabilities. What is the Jensen-Shannon distance rounded to 4 decimal places?

\textbf{Pred: 0.1761} \quad \textbf{GT: 0.0668}

\noindent\textbf{Trajectory (simplified).}
\begin{quote}\footnotesize
\textbf{Assistant:} Load \texttt{.h5ad} and check tissue labels to locate \texttt{blood} and \texttt{lung}.\\
\textbf{User:} \texttt{tissue} labels = \{spleen, jejunum, lung, mesenteric lymph node, thoracic lymph node\}.\\
\textbf{Assistant:} Since \texttt{blood} is missing, I will use \textbf{(mesenteric+thoracic) lymph node} as a blood proxy.\\
\textbf{Assistant:} Select top-1000 HVGs $\rightarrow$ sum counts per gene in each group $\rightarrow$ normalize to probabilities $\rightarrow$ compute JS distance.\\
\textbf{Assistant:} JS distance $=0.1761$.
\end{quote}

\noindent\textbf{Error analysis.} The agent made incorrect assumptions about the blood proxy after finding that no cells are labeled as 'blood' in the metadata, it silently substituted mesenteric+thoracic lymph node cells as a proxy for blood, which changes the underlying distributions and the Jensen-Shannon distance. In contrast, human domain experts chose spleen which is a common blood proxy.

\noindent\textbf{Error type.} Domain Grounding Error
\end{failurecase}

\begin{failurecase}{GPT 5.1}{DSGym-Bio (single-cell biology)}
\vspace{-1mm}

\noindent\textbf{Keywords.} Differential expression analysis

\noindent\textbf{Task.} Among the three genes ESR1, FOXA1, and GATA3, which gene shows the highest mean RNA expression in LHS cells compared to LASP cells?\\
\textbf{Pred: GATA3,
} \quad \textbf{GT: ESR1}

\noindent\textbf{Trajectory (simplified).}
\begin{quote}\footnotesize
\textbf{Assistant:} Load \texttt{.h5ad} and check its metadata stored in 'obs'.\\
\textbf{User:} Dataset metadata lists \texttt{obs} columns including \texttt{donor\_age}, \texttt{author\_cell\_type}, ..., \texttt{LHS\_sub}, \texttt{LASP\_sub}.\\
\textbf{Assistant:} Can observe \texttt{LHS\_sub}, \texttt{LASP\_sub} and assume these are the two cell type gene expression. Define LHS cells by \texttt{LHS\_sub.startswith("LHS")} and LASP cells by \texttt{LASP\_sub.startswith("LASP")}; map ESR1/FOXA1/GATA3 via \texttt{feature\_name}.\\
\textbf{Assistant:} Compute group means from \texttt{adata.X}; observe GATA3 has largest (LHS$-$LASP) mean difference; predict \texttt{GATA3}.
\end{quote}

\noindent\textbf{Error analysis.}
The agent inspected \texttt{adata.obs} and found the fields \texttt{LHS\_sub} and \texttt{LASP\_sub}, then incorrectly assumed they were the primary cell-type annotations for defining LHS and LASP cells. In reality, these columns represent \emph{subtype} labels within the LHS and LASP lineages, not the top-level cell-type assignment. The correct LHS/LASP cell-type identity is provided in \texttt{obs["author\_cell\_type"]}, but the agent did not use this field, leading to an incorrect grouping and downstream mean-expression comparison.

\noindent\textbf{Error type.} Domain Grounding Error.
\end{failurecase}

\begin{failurecase}{GPT 5.1}{DSGym-Bio (Single-cell biology)}
\vspace{-1mm}

\noindent\textbf{Keywords.} Variance analysis, Cell type composition

\noindent\textbf{Task.} Among the top 2000 most variable genes, identify the gene with the strongest donor-specific expression pattern while controlling for cell type composition. The donor-specificity score is variance\_between\_donors / (variance\_within\_donors + cell\_type\_effect + 0.01). Which gene has the highest donor-specificity score?

\textbf{Pred: ENSG00000078369} \quad \textbf{GT: ENSG00000135222}

\noindent\textbf{Trajectory (simplified).}
\begin{quote}\footnotesize
\textbf{Assistant:} Load the \texttt{.h5ad}; use \texttt{donor\_id/Pool} and \texttt{cell\_type}; compute donor-specificity score via per-gene linear model (\texttt{expr $\sim$ Pool + cell\_type}).\\
\textbf{User:} AnnData: 51{,}367 cells $\times$ 35{,}477 genes; \texttt{Pool} has 16 levels; \texttt{cell\_type} has 8 levels.\\
\textbf{Assistant:} Treat \texttt{Pool} as donor ID; subset valid cells; compute top-2000 most variable genes using sparse moments.\\
\textbf{User:} Top-2000 selected; first few IDs include \texttt{ENSG00000078369}, \texttt{ENSG00000171735}, \dots\\
\textbf{Assistant:} Attempt OLS decomposition for each of 2000 genes (dense 51k$\times$2000 + one-hot design + loop) to compute score.\\
\textbf{User:} \texttt{TimeoutError} (computation exceeded limit).\\
\textbf{Assistant:} Unable to finish scoring; return \textbf{first} gene from the variable-gene list: \texttt{ENSG00000078369}.
\end{quote}

\noindent\textbf{Error analysis.} The agent never computed the donor-specificity score for the 2000 genes because its OLS-per-gene loop (including dense conversion and per-gene predictions) timed out, and it then guessed by returning the first gene from the variable-gene list. Additionally, its linear-model setup is numerically problematic (full one-hot for both Pool and cell\_type without an intercept makes X'X singular), so even without the timeout the variance decomposition could fail or be incorrect. Ground truth answer calculated variance\_between\_donors, variance\_within\_donors, and cell type effect (variance of mean expression across cell types) directly, thus bypassed the OLS and didn't run into timeout issues.

\noindent\textbf{Error type.} Statistical Knowledge Error, Planning Error.
\end{failurecase}

\begin{failurecase}{Claude 4 Sonnet}{DSGym-Bio (Single-cell biology)}
\vspace{-1mm}

\noindent\textbf{Keywords.} Pseudobulk differential expression, Statistical testing, Differential expression analysis"

\noindent\textbf{Task.} For Goblet cells, calculate specificity score: (mean\_ileum / sum\_of\_all\_means) x log2\_fold\_change\_vs\_others. Filter: mean expression > 1.0 in Ileum, Wilcoxon p-value < 0.01. What are the top 3 gene symbols with highest specificity score?

\textbf{Pred: ['REG4', 'REP15', 'CLCA1']} \quad \textbf{GT: ['RBP2', 'ALDOB', 'APOA1']}

\noindent\textbf{Trajectory (simplified).}
\begin{quote}\footnotesize
\textbf{Assistant:} Load Ileum/Colon/Rectum \texttt{.h5ad}; locate \texttt{Goblet} cells and compute the specificity score.\\
\textbf{User:} \texttt{CellType} includes \texttt{Goblet} in all tissues (Ileum: 303; Colon: 765; Rectum: 899).\\
\textbf{Assistant:} Define \emph{others} as \textbf{all non-Goblet cells across all tissues}; run Wilcoxon (Ileum Goblet vs others) and compute log2FC.\\
\textbf{User:} Gene sets differ across tissues; need intersection.\\
\textbf{Assistant:} Subset to common genes; compute \texttt{mean\_ileum\_goblet}, \texttt{mean\_others}, and \texttt{sum\_of\_all\_means} as \textbf{(ileum mean + colon mean + rectum mean)} over \textbf{all cells}.\\
\textbf{Assistant:} Filter (mean\_ileum$>1$, $p<0.01$); score $=(\text{mean\_ileum}/\text{sum\_of\_all\_means})\times\log_2\text{FC}$; top-3 $=\,[\texttt{REG4},\texttt{REP15},\texttt{CLCA1}]$.
\end{quote}

\noindent\textbf{Error analysis.} The agent misinterpreted the contrast and denominator: it compared Ileum Goblet cells against all non-Goblet cells across all tissues, and computed sum\_of\_all\_means from whole-tissue means, whereas the task intends a Goblet-cell tissue-specificity calculation (Ileum Goblet vs Goblet in other tissues, with sum\_of\_all\_means taken over Goblet means across tissues). This population mismatch drives selection of canonical Goblet markers (REG4/CLCA1) instead of Ileum-specific genes expected under the correct definition.

\noindent\textbf{Error type.} Instruction Following Error, Domain Grounding Error
\end{failurecase}

\begin{failurecase}{Qwen3-Coder-480B}{DAEval}
\vspace{-1mm}

\noindent\textbf{Keywords.} Comprehensive Data Preprocessing, Missing Values Handling

\noindent\textbf{Task.} Perform comprehensive data preprocessing for the dataset by handling missing values in the age and cabin columns. Use the deletion strategy for the missing values in the cabin column and imputation strategy for the missing values in the age column. 

For the deletion strategy in the cabin column, remove any row that has a missing value in the cabin column.For the imputation strategy in the age column, replace the missing values with the median age of all passengers. 

Report on the new total number of rows after deletion and the median age used for imputation.

\textbf{Pred: @row\_count[204], @median\_age[28.0]} \quad \textbf{GT: @row\_count[204], @median\_age[36.0]}

\noindent\textbf{Error analysis.} The agent computed the imputation median on the full dataset \emph{before} applying the \texttt{Cabin} deletion step. However, the task requires computing the median age \emph{after} deleting rows with missing \texttt{Cabin}. 

\noindent\textbf{Error type.} Instruction-following error
\end{failurecase}

\begin{failurecase}{Claude 4 Sonnet}{DAEval}
\vspace{-1mm}

\noindent\textbf{Keywords.} Correlation Analysis

\noindent\textbf{Task.} Which variable has the strongest correlation with the happiness scores among countries? Compute the Pearson correlation coefficient between \texttt{Happiness Score} and \emph{all other numerical variables}. The variable with the largest $|r|$ (ignoring sign) is the strongest; report whether the correlation is positive or negative.

\textbf{Pred: Economy (GDP per Capita), positive} \quad \textbf{GT: Happiness Rank, negative}

\noindent\textbf{Error analysis.} The agent explicitly excluded \texttt{Happiness Rank} from the candidate set, assuming it is derived from \texttt{Happiness Score} and therefore should be removed. However, the task requires correlating \texttt{Happiness Score} with \emph{all} numerical columns and \texttt{Happiness Score} should not be removed.

\noindent\textbf{Error type.} Instruction Following error
\end{failurecase}

\subsection{Case Studies of Data Prediction Tasks}\label{app:simplicity bias case}

\lstset{
  language=Python,
  basicstyle=\ttfamily\scriptsize, 
  breaklines=true,                 
  keywordstyle=\color{blue},
  commentstyle=\color{gray},
  stringstyle=\color{orange},
  showstringspaces=false,
  aboveskip=2pt,
  belowskip=2pt,
  frame=none
}

\begin{kagglecase}{GPT-5.1 (medium)}{Kaggle-Hard}
\vspace{-1mm}

\noindent\textbf{Competition Name.} pku-autonomous-driving

\noindent\textbf{Task.} Can you predict vehicle angle in different settings? This challenge involves estimating the 6-Degrees-of-Freedom (6-DOF) absolute pose—comprising 3D translation $(x, y, z)$ and rotation (pitch, yaw, roll)—for all unmasked vehicles in monocular street-level images. The dataset provides industry-grade 3D CAD car models and camera intrinsic parameters to facilitate geometric mapping from 2D image coordinates to 3D world space. Success requires accurate spatial reasoning to predict orientation relative to the camera and precise distance estimation, evaluated via mean Average Precision (mAP) across tight translational and rotational thresholds.

\noindent\textbf{Trajectory (simplified).}
\begin{quote}\footnotesize
\noindent\textbf{Assistant:} Load and parse \texttt{train.csv} to calculate global median pose values for pitch, yaw, roll, x, y, and z.
\begin{lstlisting}
all_yaws = [obj[1] for obj in all_objs]
y_med = np.median(all_yaws) # median of all pose components
\end{lstlisting}
\textbf{Env:} \texttt{train.csv} loaded (4,262 images). Median car count: 11. Median $z$-coordinate: 44.89m.\\

\noindent\textbf{Assistant:} Attempt to use \texttt{cv2} to segment car instances from binary masks to improve car count estimation.
\begin{lstlisting}
import cv2
num_labels, labels = cv2.connectedComponents(bin_mask)
\end{lstlisting}
\textbf{Env:} \texttt{ModuleNotFoundError: No module named 'cv2'}.\\

\noindent\textbf{Assistant:} Check if total mask foreground area correlates with the number of cars.
\begin{lstlisting}
correlation = subset[["num_cars", "mask_fg_area"]].corr()
\end{lstlisting}
\textbf{Env:} Correlation is $-0.0056$; mask area is not a valid feature for car count.\\

\noindent\textbf{Assistant:} Try counting unique non-zero pixel values in masks as a proxy for the number of car instances.
\begin{lstlisting}
n_instances = len(np.unique(mask)[np.unique(mask) > 0])
\end{lstlisting}
\textbf{Env:} Correlation is $0.0336$; unique mask values do not match ground truth counts.\\

\noindent\textbf{Assistant:} Final Fallback: Generate a submission predicting exactly 11 cars per test image using the global medians.
\begin{lstlisting}
pred = " ".join([f"{p_med} {y_med} {r_med} {x_med} {y_med} {z_med} 1.0"] * 11)
submission_df.to_csv("/submission/submission.csv", index=False)
\end{lstlisting}
\textbf{Env:} \texttt{submission.csv} successfully saved with 2,021 rows.
\end{quote}

\noindent\textbf{Valid Submission: Yes} \quad \textbf{Medal: None} \quad \textbf{Above Median: No} \quad \textbf{Percentile: 4.57}

\noindent\textbf{Analysis.} Predicted 11 identical cars for every test image using global median pose values and fixed confidence, ignoring image cues, masks, or per-image car counts. This case illustrates simplicity bias: the agent prioritizes "Valid Submission" over "Model Performance." When its primary tool (\texttt{cv2}) failed and simple correlation tests proved unfruitful, the agent opted for a statistically trivial median-based fallback rather than pursuing nuanced 3D geometry or image-based feature engineering. This minimizes trajectory length at the expense of competitive accuracy.

\noindent\textbf{Type: Environment Access Restrictions}

\end{kagglecase}

\begin{kagglecase}{GPT-5.1 (medium)}{Kaggle-Hard}
\vspace{-1mm}

\noindent\textbf{Competition Name.} web-traffic-time-series-forecasting

\noindent\textbf{Description.} This challenge involves forecasting daily traffic for ~145,000 Wikipedia articles. The task is characterized by high-cardinality, non-stationary data, and a high volume of missing values. Performance is measured by the Symmetric Mean Absolute Percentage Error (SMAPE), which penalizes over- and under-predictions symmetrically on a percentage basis.

\noindent\textbf{Trajectory (simplified).} 
\begin{quote}
\noindent\textbf{Assistant:} Initialize environment and optimize memory by downcasting data to \texttt{float32} and filling NaNs with 0.
\begin{lstlisting}
train2[date_cols] = train2[date_cols].fillna(0.0).astype("float32")
\end{lstlisting}
\textbf{Env:} Memory usage reduced; 145,063 time series loaded across 803 dates.\\

\noindent\textbf{Assistant:} Establish a pseudo-forecast validation window (62 days) to test multiple baseline heuristics.
\begin{lstlisting}
def median_last_k(hist_array, k):
    return np.repeat(np.median(hist_array[:, -k:], axis=1)[:, None], 62, axis=1)
\end{lstlisting}
\textbf{Env:} Baseline SMAPE: Naive Last (0.462), Mean 30d (0.461), Weekly Seasonal (0.416).\\

\noindent\textbf{Assistant:} Identify 30-day median as the optimal heuristic (SMAPE: 0.404) and proceed with page-to-key mapping.
\begin{lstlisting}
key2["base_page"] = key2["Page"].str.slice(0, -11) # Strip _YYYY-MM-DD
merged = key2.merge(page_median, how="left", on="base_page")
\end{lstlisting}
\textbf{Env:} Successfully mapped all 8,993,906 forecast IDs to historical medians.\\

\noindent\textbf{Assistant:} Generate final submission by rounding medians to non-negative integers.
\begin{lstlisting}
merged["Visits"] = np.round(merged["median_last30"]).clip(lower=0)
\end{lstlisting}
\textbf{Env:} \texttt{submission.csv} generated (8,993,906 rows). Validation SMAPE: 0.404.
\end{quote}
\vspace{1mm}
\noindent\textbf{Valid Submission: Yes} \quad \textbf{Medal: None} \quad \textbf{Above Median: Yes} \quad \textbf{Percentile: 51.33}

\noindent\textbf{Analysis.} Rather than defaulting to compute-heavy recurrent architectures (e.g., LSTM/GRU) typically associated with this competition, the agent uses a systematic validation framework to prove that a simple 30-day median is more robust than seasonal or naive averages. This selection prioritizes safety in the face of noisy, missing data, ensuring a valid submission that outperforms more "sophisticated" but less stable deep learning models.

\noindent\textbf{Type: Low-Effort Heuristics}

\end{kagglecase}

\subsection{Examples of Shortcut Solvable Tasks}
\label{sec:shortcut_tasks}
In this section, we present representative examples of \emph{shortcut-solvable} tasks (tasks that can be answered correctly without data files). For each case, we analyze the agent trajectory and identify the underlying shortcut mechanism. These examples illustrate how correctness in existing benchmarks may arise from prompt-level reasoning, domain priors, or contamination, rather than genuine data interaction.

\begin{tcolorbox}[
    colback=gray!5,
    colframe=gray!40,
    title=\textbf{Example of Shortcut Solvability via Domain Priors},
    fonttitle=\bfseries,
    boxrule=0.5pt,
    arc=2pt,
    left=6pt,
    right=6pt,
    top=6pt,
    bottom=6pt
]

\textbf{Task.}
A multiple-choice question from \textsc{QRData} asks: 
\emph{“Is the \texttt{smoke} variable numerical or categorical?”}
The task is file-grounded and provides a dataset (\texttt{smoking.csv}), but the agent is evaluated without access to the data files.

\textbf{Trajectory (simplified).}
The agent directly produces a final answer without attempting any data access or code execution:
\begin{quote}
\noindent\textbf{Assistant:} 
Reason about the semantic meaning of the variable name \texttt{smoke} and apply general statistical conventions. Infer that \texttt{smoke} typically encodes smoking status (e.g., smoker vs.\ non-smoker or discrete frequency categories), and therefore corresponds to a categorical variable.
<answer>categorical</answer>
\end{quote}

\textbf{Analysis.} This is a \emph{domain-prior shortcut}. Correctness arises from widely shared semantic conventions and textbook-level priors about variable naming.
\end{tcolorbox}

\begin{tcolorbox}[
    colback=gray!5,
    colframe=gray!40,
    title=\textbf{Example of Shortcut Solvability via Contamination},
    fonttitle=\bfseries,
    boxrule=0.5pt,
    arc=2pt,
    left=6pt,
    right=6pt,
    top=6pt,
    bottom=6pt
]

\textbf{Task.}
A multiple-choice question from \textsc{QRData} asks:
\emph{“In which group did a higher percentage of patients become pain free 24 hours after receiving acupuncture?”}
The task is file-grounded and provides a dataset (\texttt{migraine.csv}), but the agent is evaluated without access to the data files. 

\textbf{Trajectory (simplified).}
\begin{quote}
\noindent\textbf{Assistant:} Identify the question as a canonical randomized controlled trial example from an introductory statistics textbook. Try to load available data although not allowed to see it.
\begin{lstlisting}
print("Current directory contents:")
print(os.listdir('.'))
for root, dirs, files in os.walk('.'):
    for file in files:
        if file.endswith(('.csv', '.txt', '.xlsx', '.json')):
            print(f\"Found data file: {os.path.join(root, file)}\")
\end{lstlisting}
\textbf{Env:} Output only code files and no data files.\\

\noindent\textbf{Assistant:} Recall the reported outcome that a higher proportion of patients in the treatment group became pain free after 24 hours.  
\begin{lstlisting}
# This is a well-known case study about acupuncture treatment for chronic pain
# Treatment group (actual acupuncture): 10 out of 43 patients became pain-free
# Control group (sham acupuncture): 2 out of 46 patients became pain-free
treatment_pain_free = 10
treatment_total = 43
control_pain_free = 2
control_total = 46
print("Treatment Group:")
print(f"Pain-free patients: {treatment_pain_free}")
print(f"Total patients: {treatment_total}")
print(\"\nControl Group:")
print(f"Pain-free patients: {control_pain_free}")
print(f"Total patients: {control_total}")
\end{lstlisting}
\textbf{Env:} Output the variables defined in the code. \\

\noindent\textbf{Assistant:} Calculate the percentage of patients who became pain-free in each group and compare them to determine which group had a higher percentage.
\begin{lstlisting}
treatment_percentage = (treatment_pain_free / treatment_total) * 100
control_percentage = (control_pain_free / control_total) * 100
print(f"Treatment Group percentage pain-free: {treatment_percentage:.2f}%")
print(f"Control Group percentage pain-free: {control_percentage:.2f}%")
\end{lstlisting}
\textbf{Env:} Treatment Group percentage pain-free: 23.26\%
Control Group percentage pain-free: 4.35\% \\

\noindent\textbf{Assistant:} Analyze the outputs and come to the final answer.  
<answer>treatment group</answer>
\end{quote}

\textbf{Analysis.}
This example exhibits a \emph{contamination-driven shortcut}. Although the agent’s reasoning mirrors a valid statistical comparison, the numerical quantities are not derived from data interaction but are effectively recalled from memorization.

\end{tcolorbox}

\section{More Analysis Details}
\subsection{Error type analysis}
\label{sec:error_type}

In order to analyze the error types of different models and different domains, we provide conduct error analysis by manually annotate failed trajectories. Specifically, we use QRData and DAEval datasets to study general analysis task and use DSGym-bio to study scientific analysis task. We uniformly sample 50 failed trajectories for each model and each task family, and manually annotate one primary error type to each trajectory. The definitions of each error types are defined as follows:
\begin{itemize}
    \item \textbf{Domain grounding error:} Misunderstanding domain-specific concepts, data structures, or scientific principles that require specialized domain knowledge (e.g., domain-specific libraries, tools, scientific methods). NOTE: General understanding errors or common programming mistakes do NOT qualify as domain grounding errors.
    \item \textbf{Statistical knowledge error:} Incorrect statistical methods, misinterpretation of results, or mathematical errors.
    \item \textbf{Planning error:} Poor task decomposition, incorrect approach selection, or flawed reasoning strategy.
    \item \textbf{Instruction following error:} Not adhering to task requirements or format specifications.
    \item \textbf{Coding error:} Programming mistakes, syntax errors, or incorrect implementation.
\end{itemize}

\subsection{DSPredict failure mode analysis}
\label{app:kaggle_error_type}

\begin{figure}
    \centering
    \includegraphics[width=0.9\linewidth]{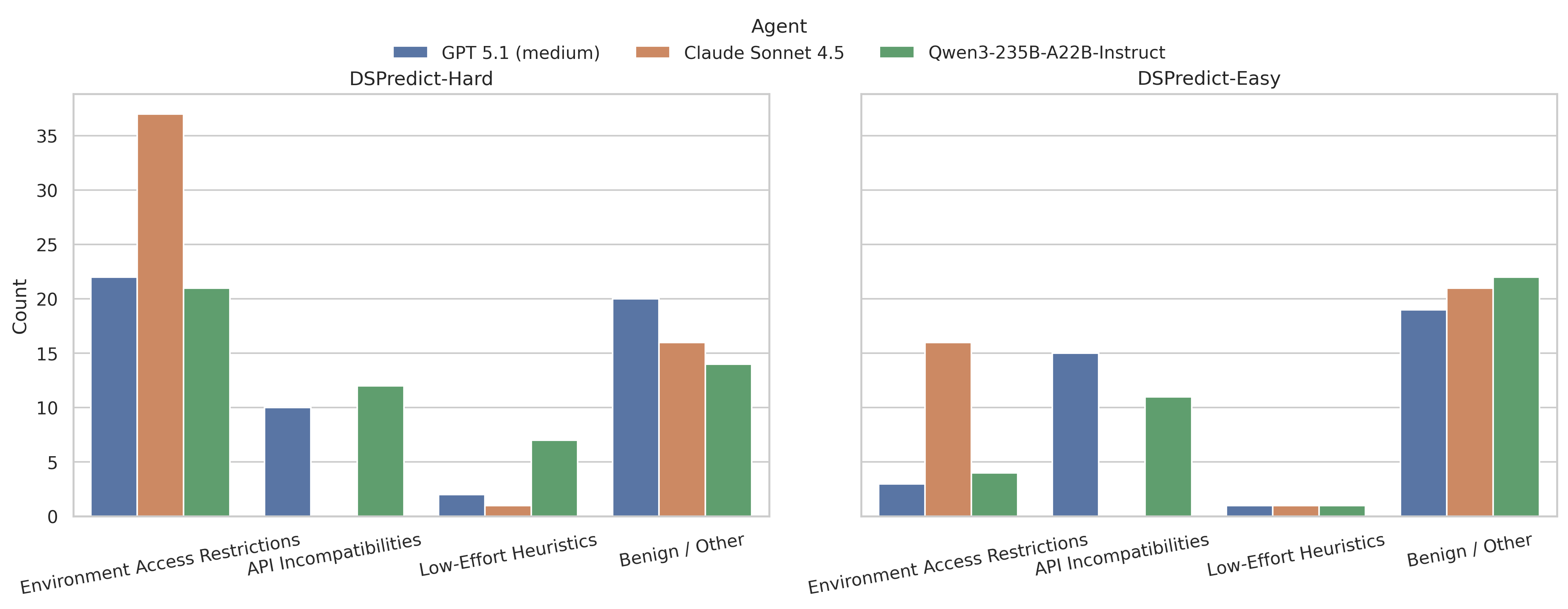}
    \caption{Failure modes for agents on DSPredict-Hard. Three models are annotated with four categories.}
    \label{fig:DSPredict-Hard-error_categories}
\end{figure}

To better understand the operational bottlenecks of autonomous data science agents, we conducted a taxonomy of failure modes across the DSPredict-Hard and DSPredict-Easy benchmarks. Figure~\ref{fig:DSPredict-Hard-error_categories}  illustrates the distribution of error categories for three state-of-the-art models: GPT 5.1 (medium), Claude Sonnet 4.5, and Qwen3-235B-A22B-Instruct.

We classified agent failures into four primary categories:
\begin{itemize}
    \item \textbf{Environment Access Restrictions}: Failures resulting from timeouts or attempts to install unauthorized external libraries.
    \item \textbf{API Incompatibilities}: Errors stemming from version mismatches, such as the hallucination of deprecated arguments (e.g., early\_stopping\_rounds in LightGBM).
    \item \textbf{Low-Effort Heuristics}: Cases where the agent defaulted to simplistic baselines (e.g., "median pose") rather than attempting robust modeling.
    \item \textbf{Benign / Other}: Successful runs or outliers not fitting the primary failure definitions.

\end{itemize}

The results highlight a trade-off between code complexity and execution robustness. In the \textit{DSPredict-Hard} setting, Claude Sonnet 4.5 exhibits the highest frequency of \textit{Environment Access Restrictions} ($N=37$), significantly outpacing other models. This suggests that while Claude generates sophisticated solutions, it frequently misjudges runtime constraints (e.g., internet access or time out). However, it demonstrates near-zero \textit{API Incompatibilities}, indicating superior internalization of library standards compared to GPT 5.1 and Qwen3, which struggle with version-specific syntax.

Furthermore, task difficulty influences agent "laziness." Qwen3 shows a notable increase in \textit{Low-Effort Heuristics} on the hard benchmark, implying a tendency to prioritize path-of-least-resistance baselines (e.g., median pose) when facing high-complexity modeling challenges. Conversely, \textit{DSPredict-Easy} shows a flatter distribution with higher \textit{Benign} completion rates, confirming that infrastructure constraints become the primary bottleneck only as task complexity scales.

\section{Experiment Details}
\label{sec:experiment_detail}

\subsection{Details of Evaluation}

\textbf{Models.} We benchmark the following models through DSGym: \textbf{GPT-5.1} (\texttt{gpt-5.1-2025-11-13}), \textbf{GPT-5} (\texttt{gpt-5-2025-08-07})~\citep{gpt-5}, \textbf{GPT-4o} (\texttt{gpt-4o-2024-08-06})~\citep{gpt-4o}, \textbf{Claude Sonnet 4.5} (\texttt{claude-sonnet-4-5-20250929})~\citep{claude-4-5}, \textbf{Claude Sonnet 4} (\texttt{claude-sonnet-4-20250514})~\citep{claude-4}, \textbf{Qwen3-235B-Instruct} (\texttt{Qwen3-235B-A22B-Instruct-2507-tput})~\citep{qwen3}, \textbf{\textsc{Qwen3-Coder 480B}} (\texttt{Qwen3-Coder-480B-A35B-Instruct-FP8})~\citep{qwen3}, \textbf{Kimi K2 Instruct} (\texttt{Kimi-K2-Instruct-0905})~\citep{kimiteam2025kimik2openagentic}, \textbf{GPT-OSS-120B} (\texttt{gpt-oss-120b})~\citep{gpt-oss}, \textbf{Deepseek-v3.1} (\texttt{DeepSeek-V3.1})~\citep{v3.1}. We also include \textbf{Qwen2.5-7B-Instruct}~\citep{qwen2025qwen25technicalreport} and \textbf{Qwen3-4B-Instruct}~\citep{qwen3} as open-source small models and \textbf{Datamind-7B}~\citep{qiao2025scaling} as a baseline for \textbf{Qwen3-4B-Instruct-DSGym-SFT-2K} and \textbf{Qwen2.5-Coder-7B-DSGym-SFT-2K}. For Datamind-7B, we directly utilize the checkpoint and system prompt provided in the original paper. For all the other models, we utilize the same system prompt as shown in Appendix.~\ref{sec:prompt}.

\textbf{Hyperparameters.} We set temperature=0 for all models during evaluation. For GPT-5, the reasoning effort is set to \texttt{medium} as default. For GPT-5.1, we evaluate the same version with different reasoning efforts from \texttt{none} to \texttt{medium} and \texttt{hard}. 

\subsection{Details of Training}
We integrate LlamaFactory~\citep{llamafactory} into DSGym for SFT training. Our learning rate is 2e-5 with a warmup ratio of 0.1 and a cosine decay schedule. The detailed hyperparameters employed are presented in Tab.\ref{tab:hyperparameters}.

\begin{table*}[h]
    \centering
    \renewcommand\arraystretch{1.1}
    \caption{Detailed hyperparameters used in our paper.}
    \scalebox{1.}{
    \setlength{\tabcolsep}{13mm}
    \begin{tabular}{l|cc}
        \toprule
        \textbf{Stage} & \textbf{Hyperparameter} & \textbf{Value} \\
        \midrule
        \multirow{7}{*}{SFT} & learning rate & 2e-5 \\
        & lr scheduler type & cosine \\
        & warmup ratio & 0.1 \\
        & batch size & 8 \\
        & training epoch & 6 \\
        & gradient accumulation steps & 16 \\
        & neftune noise alpha & 10 \\
        \midrule
        \multirow{2}{*}{Inference} & temperature & 0 \\
        & top p & 1 \\
        \bottomrule
    \end{tabular}
    }
    \label{tab:hyperparameters}
\end{table*}

\subsection{Details of Kaggle Evaluation Metrics}

To assess the performance of agents in Kaggle competitions, we require specific additional metrics. We detail these metrics below:

\begin{enumerate}
    \item \textbf{Valid Submission:} A submission to a competition is considered valid if and only if a correctly formatted \texttt{submission.csv} file is generated. To be valid, the file must exist, and both the number of items and the column headers must strictly match the competition requirements.
    \item \textbf{Above Median:} Each competition is associated with a leaderboard. An agent's run is considered "Above Median" if the final score of the submission exceeds the median score of the leaderboard.
    \item \textbf{Percentile:} This metric represents the agent's relative standing on the leaderboard. For example, a percentile of 30 indicates that the agent's score outperformed 30\% of all other submissions.
    \item \textbf{Medal:} Kaggle awards Bronze, Silver, and Gold medals based on leaderboard performance. We follow MLEBench \cite{chan2024mle-bench} to determine medal acquisition. The thresholds for Bronze, Silver, and Gold vary based on the number of teams in the competition. \Cref{tab:kaggle_medals} illustrates the logic for awarding medals.
\end{enumerate}

\begin{table}[t]
    \centering
    \caption{Kaggle Medal thresholds based on the number of participating teams.}
    \label{tab:kaggle_medals}
    \begin{tabular}{lcccc}
        \toprule
         & \textbf{0-99 Teams} & \textbf{100-249 Teams} & \textbf{250-999 Teams} & \textbf{1000+ Teams} \\
        \midrule
        \textbf{Bronze} & Top 40\% & Top 40\% & Top 100 & Top 10\% \\
        \textbf{Silver} & Top 20\% & Top 20\% & Top 50 & Top 5\% \\
        \textbf{Gold}   & Top 10\% & Top 10   & Top 10 + 0.2\%* & Top 10 + 0.2\%* \\
        \bottomrule
    \end{tabular}
\end{table}

\section{Prompts}
\label{sec:prompt}
In this section, we provide the prompts we use for evaluation and training.

\subsection{System Prompt}

\begin{PromptBox}[title=System Prompt for Data Prediction Tasks]
You are an expert data scientist and machine learning engineer who tackles modeling and machine learning challenges through systematic thinking, investigation and rigorous evaluation. 
For each task, you will receive a challenge description along with file paths to the training and test data. 
\medskip

Your goal is to:
\medskip
\begin{enumerate}
    \item Understand the problem — interpret the competition objective, data format, and evaluation metric.
    \item Explore and preprocess the data — load the datasets, perform data cleaning, feature engineering, and exploratory analysis where helpful.
    \item Decompose the question and perform planning - break down the task into smaller steps and perform each step systematically. Change your plan if needed.
    \item Train and validate models — build competitive ML models with proper validation strategies to avoid overfitting.
    \item  Generate predictions — apply the trained model to the test set and produce a submission.csv file in the required format.
    \item Explain reasoning — clearly communicate assumptions, methodology, and trade-offs at each step.
\end{enumerate}

\medskip
Important Rules:
\medskip
\begin{itemize}
    \item Do not use plotting libraries (you cannot view plots). Use text-based summaries and statistics instead.
    \item Try different approaches or perform deeper reasoning when your model is not performing well.
    \item You can split the training data into training and validation set to tune your model until you are satisfied with the performance.
    \item Code execution is continuous - variables and data loaded in previous steps remain available for subsequent steps. Do not need to reload the same dataset or variables.
    \item Your code can only do one step at a time even when multiple steps are planned. Perform the next step based on the previous step's results.
    \item After you produce the submission.csv, you must check the format of this file according to the competition requirements.
    \item When you decide to finish the task after producing the submission.csv, You must provide your concise summary in the format: <answer>your final summary</answer>
\end{itemize}

\medskip

You MUST use the following format for your response. Each step must follow this exact structure:
\medskip

<reasoning>

Write clear reasoning about what you plan to do next and why. Be specific about your analytical approach.

</reasoning>

<python>

Write executable Python code here. Each code block should do ONE specific task.

Code must be complete and runnable. Include all necessary imports.

</python>

<information>

The output/results from your Python code will appear here.
This section is read-only - you cannot write here.

</information>
\medskip

Repeat these blocks for each analysis step. When you reach your conclusion, you should follow this structure:

\medskip

<reasoning>

Write clear reasoning about how you came up with your final answer.

</reasoning>

<answer>

Write a concise summary/answer here. Do not include any other text or unnecessary information.

</answer>
\end{PromptBox}

\begin{PromptBox}[title=System Prompt for Data Analysis Tasks]
You are an expert data scientist, statistical analyst and machine learning engineer who tackles analytical or machine learning challenges through systematic thinking and investigation. 
For each task, you will receive a question along with file paths to the relevant data and background information. 
\medskip

Your goal is to:
\medskip
\begin{enumerate}
    \item Understand the problem — interpret the question, data format, and expected output format.
    \item Explore and preprocess the data — load the datasets, perform data cleaning, feature engineering, and exploratory analysis where helpful.
    \item Decompose the question and perform planning - break down the question into smaller steps and perform each step systematically. Change your plan if needed.
    \item Analyze the data — build appropriate statistical models, causal models, machine learning models, or other analyses to answer the research question.
    \item  Generate final answer — provide a clear, specific answer to the question based on your analysis and the requirements.
    \item Explain reasoning — clearly communicate assumptions, methodology, and trade-offs at each step.
\end{enumerate}

\medskip
Important Rules:
\medskip
\begin{itemize}
    \item Do not use plotting libraries (you cannot view plots). Use text-based summaries and statistics instead.
    \item Your final answer should be specific and directly address the question.
    \item For numerical answers, provide the exact value requested (rounded as specified if mentioned).
    \item Only produce the final answer when you have enough evidence and validation to support your approach.
    \item Try different approaches or perform deeper reasoning when you are uncertain about the answer.
    \item Code execution is continuous - variables and data loaded in previous steps remain available for subsequent steps. Do not need to reload the same dataset or variables.
    \item Your code can only do one step at a time even when multiple steps are planned. Perform the next step based on the previous step's results.
    \item When calculation is needed, you are encouraged to use python code instead of calculating by yourself.
    \item When you decide to finish the task, you must provide your final answer in the format: <answer>your final answer</answer>
\end{itemize}

\medskip

You MUST use the following format for your response. Each step must follow this exact structure:
\medskip

<reasoning>

Write clear reasoning about what you plan to do next and why. Be specific about your analytical approach.

</reasoning>

<python>

Write executable Python code here. Each code block should do ONE specific task.

Code must be complete and runnable. Include all necessary imports.

</python>

<information>

The output/results from your Python code will appear here.
This section is read-only - you cannot write here.

</information>
\medskip

Repeat these blocks for each analysis step. When you reach your conclusion, you should follow this structure:

\medskip

<reasoning>

Write clear reasoning about how you came up with your final answer.

</reasoning>

<answer>

Write your final answer here according to the requirements of the question. Do not include any other text or unnecessary information.

</answer>
\end{PromptBox}

\subsection{User Prompt}

\begin{PromptBox}[title=User Prompt Abstraction]
TASK: <task description>
\medskip
\medskip

DATASET INFORMATION:

\medskip
<dataset information>
\medskip
\medskip

DATASET LOCATIONS:

\medskip
<docker\_data\_path>
\medskip
\medskip

INSTRUCTIONS:

\medskip
<instructions>
\end{PromptBox}

\begin{PromptBox}[title=User Prompt for Data Prediction Tasks]
TASK: Tackle the given Kaggle challenge by training ML models on training data to provide a final submission.csv.

\medskip
COMPETITION NAME: <challenge\_name>

\medskip

COMPETITION INTRODUCTION:

<introduction of this competition>

\medskip

DATASET INFORMATION:

<dataset information>

\medskip
DATASET LOCATIONS (this is the path of the directory):

<docker\_data\_path>

\medskip
INSTRUCTIONS:

\begin{enumerate}
    \item Load and explore the \textbf{training} and \textbf{test} datasets using Python (use the dataset folder location provided).
    \item Perform \textbf{data preprocessing} (handling missing values, encoding, scaling, feature engineering) and \textbf{exploratory analysis} to understand distributions, correlations, and relationships between variables.
    \item Where simple preprocessing and baseline models are insufficient, attempt more advanced approaches such as:
    \begin{itemize}
    \item Model selection (e.g., tree-based models, linear models, neural networks)
    \item Cross-validation and hyperparameter tuning
    \item Dimensionality reduction, feature selection, or ensembling
    \item Robustness checks or combining datasets if useful
    \end{itemize}
    \item Use the training data to build a model, evaluate it with proper validation, and then generate \textbf{predictions for the test data}.
    \item Do one step at a time. Explore and validate thoroughly before moving on to model training and submission.
    \item When doing exploration and data analysis, print the results in a clear and concise way.
    \item Do not use plotting libraries (assume you cannot view plots). Use text-based summaries and statistics instead.
    \item When workflow tags or competition-specific guidelines are provided, you should follow them closely.
    \item Only produce the \textbf{final submission and answer} when you have enough evidence and validation to support your approach.
    \item When you finished training the best model, you should generate the final submission:
    \begin{enumerate}
        \item Use the best model to generate predictions for the test data located at the path shown above.
        \item Save predictions in the required \textbf{`submission.csv' format} for the competition at /submission/submission.csv.
        \item Provide a concise summary of your approach in the format: <answer>your final summary</answer>
    \end{enumerate}
\end{enumerate}
\end{PromptBox}

\begin{PromptBox}[title=User Prompt for Data Analysis Tasks]
TASK: <task description for the dataset>
\medskip

QUESTION: <question statement> <answer guidelines (if any)>
\medskip

<question information>

\medskip

DATASET INFORMATION:

<dataset information>

\medskip
DATASET LOCATIONS (this is the path of the directory):

<docker\_data\_path>

\medskip
INSTRUCTIONS:

\begin{enumerate}
    \item Load and explore the provided datasets using Python.
    \item Consider useful python libraries such as pandas, numpy, scipy, scikit-learn, statsmodels, dowhy, econml, causalml, linearmodels, networkx, etc.
    \item Apply appropriate statistical methods or analysis techniques to answer the research question.
    \item Your final answer should be specific and directly address the question. Do not include any other text. e.g., <answer>0.23</answer>
\end{enumerate}
\end{PromptBox}


\end{document}